\documentclass[journal, 10pt,twocolumn]{IEEEtran}
\usepackage[utf8]{inputenc}
\usepackage{graphicx}
\graphicspath{./images/}

\usepackage{amssymb}
\usepackage[all,arc]{xy}
\usepackage{mathrsfs}
\usepackage{algorithm}
\usepackage{algorithmic}
\usepackage{amsbsy}
\usepackage{amsfonts}
\usepackage{amsmath}
\usepackage{amssymb}
\usepackage{amsthm}
\usepackage{array}
\usepackage{authblk}
\usepackage{calrsfs}

\usepackage{bbm} 
\usepackage[font=footnotesize,labelfont=bf]{caption}
\usepackage{color}
\usepackage[mathscr]{eucal}
\usepackage{mathrsfs}

\usepackage{float}
\usepackage{hyperref} 
\usepackage{stmaryrd}
\usepackage[font=scriptsize,labelfont=scriptsize]{subcaption}
\usepackage{graphicx}
\usepackage{url}
\usepackage{verbatim}
\usepackage{adjustbox}
\usepackage[usenames,dvipsnames,svgnames,table]{xcolor}
\usepackage{multirow,array}

\DeclareMathOperator*{\argmax}{arg\,max}

\title{Hyperspectral Image Clustering with Spatially-Regularized Ultrametrics}

\author{Shukun Zhang  \and ~ James M. Murphy
\IEEEcompsocitemizethanks{
\IEEEcompsocthanksitem S. Zhang is with the Departments of Computer Science and Mathematics at Tufts University; email: Shukun.Zhang@tufts.edu
\IEEEcompsocthanksitem J.M. Murphy is with the Department of Mathematics at Tufts University; email: JM.Murphy@tufts.edu
}
}

\begin{document}

\maketitle

\begin{abstract}

We propose a method for the unsupervised clustering of hyperspectral images based on spatially regularized spectral clustering with ultrametric path distances.  The proposed method efficiently combines data density and geometry to distinguish between material classes in the data, without the need for training labels.  The proposed method is efficient, with quasilinear scaling in the number of data points, and enjoys robust theoretical performance guarantees.  Extensive experiments on synthetic and real HSI data demonstrate its strong performance compared to benchmark and state-of-the-art methods.  In particular, the proposed method achieves not only excellent labeling accuracy, but also efficiently estimates the number of clusters.

\end{abstract}

\section{Introduction}

Remote sensing image processing has been revolutionized by machine learning.  When large training sets are available, supervised methods such as support vector machines, decision trees, and deep neural networks accurately label pixels in a wide range of imaging modalities.  However, it is often impractical to acquire the large training sets necessary for these methods to work well.  In these situations, it is necessary to develop \emph{unsupervised} clustering methods, which label the entire data set without the need for training data.  

While many methods for unsupervised clustering have been proposed, most only enjoy mathematical performance guarantees under very restrictive assumptions on the underlying data.  For example, $K$-means clustering works well for clusters that are roughly spherical and well-separated, but provably fails when the clusters becomes nonlinear or poorly separated \cite{Ng2001}.  Clustering methods based on deep learning may perform well in some instances, but are sensitive to metaparameters and lack robust mathematical performance guarantees even in highly idealized settings \cite{Baldi2012_Autoencoders, Song2013_Auto}.  

Recently, \emph{ultrametric path distances (UPD)} have been proven to provide state-of-the-art theoretical results for clustering high-dimensional data \cite{Little2020}.  In particular, only very weak assumptions on the shape and structure of the clusters are required, namely that the underlying data exhibits intrinsically low-dimensional structure.  This suggests UPD are well-suited for hyperspectral images (HSI) \cite{Le2020_Minimax}, which while very high-dimensional, are typically such that each class in the data depends (perhaps nonlinearly) on only a small number of latent variables, and in this sense are intrinsically low-dimensional.  

In this paper, we develop an HSI clustering algorithm based on \emph{ultrametric spectral clustering}.  Taking advantage of recent theoretical developments, UPDs are used to construct a weighted graph, from which a graph Laplacian is defined and then spatially regularized.  The lowest frequency eigenfunctions of this graph Laplacian are then used as features for $K$-means clustering.  The proposed method is denoted \emph{spatially regularized ultrametric spectral clustering (SRUSC)}.  SRUSC is computationally efficient---quasilinear in the number of data points---and has few tuneable parameters.  Moreover, the proposed method is shown to outperform a range of benchmark and state-of-the-art clustering methods on several synthetic and real HSI. In particular, the proposed method demonstrates strong clustering accuracy and efficient estimation of the number of latent clusters in the data.

The remainder of this article is organized as follows.  In Section \ref{sec:Background}, we provide background on unsupervised HSI clustering and ultrametric path distances.  In Section \ref{sec:Algorithm}, we detail the proposed algorithm and discuss its theoretical properties and complexity.  In Section \ref{sec:ExperimentalAnalysis}, we introduce several data sets and perform comparisons between the proposed method and related methods.  We conclude in Section \ref{sec:Conclusions}.  Code implementing the proposed method and all experimental results is publicly available\footnote{\url{https://github.com/ShukunZhang/Spatially-Regularized-Ultrametrics}}.

\section{Background}
\label{sec:Background}

Unsupervised clustering consists in providing a data set $X=\{x_{i}\}_{i=1}^{n}\subset\mathbb{R}^{D}$ with labels $\{y_{i}\}_{i=1}^{n}$, where each $y_{i}\in \{1,2,\dots, K\}$.  In the case of HSI, $n$ is the number of pixels in the data set, $D$ the number of spectral bands, and $K$ the number of latent classes in $X$.  The key challenge in clustering, compared to supervised learning, is that no training data is available to guide the labeling process.  So, labeling decisions must be made entirely based on latent geometrical and statistical properties of the data. 

A range of methods for clustering data have been developed, including $K$-means clustering and Gaussian mixture models, which assume the underlying data is a mixture of well-separated, roughly spherical Gaussians; density-driven methods that characterize clusters as high density regions separated from other high-density regions by regions of low density \cite{Ester1996, Rodriguez2014}; and spectral graph methods that attempt to find communities in networks generated from the underlying data \cite{Shi2000normalized,  Ng2001}.  In the specific context of clustering HSI, methods taking advantage of the intrinsically low-dimensional (though potentially nonlinear) structure of the HSI clusters are particularly important, because capturing such low-dimensional structures significantly lowers the sampling complexity necessary to defeat the curse of dimensionality.  

\subsection{Background on Ultrametic Path Distances}
\label{subsec:PathDistancesBackground}

In many clustering algorithms, decisions are based on pairwise distances between data points; the choice of distance metric is thus critical.  We propose to use UPDs.

Let $\mathcal{G}_{0}=(X,W)$ be an undirected Euclidean $k$-nearest neighbor graph on $X$, with $k\sim\log(n)$.  For any $x_{i},x_{j}\in X$, let $\mathcal{P}(x_{i},x_{j})$ be the set of paths connecting $x_{i}, x_{j}$ in $\mathcal{G}_{0}$.   Define the \emph{ultrametric path distance (UPD)} between $x_{i},x_{j}\in X$ as \begin{align}\label{eqn:UPD} \rho_{\infty}(x_{i},x_{j})=\min_{\{\gamma_{\ell}\}_{\ell=1}^{L}\in\mathcal{P}(x_{i},x_{j})}\max_{\ell=1,\dots,L-1}\|\gamma_{\ell+1}-\gamma_{\ell}\|_{2}.
\end{align}Intuitively, UPD computes the longest edge in each path, then minimizes this quantity over all paths.  It may be understood as an $\ell^{\infty}$-geodesic, while the classical shortest path is the $\ell^{1}$-geodesic.  Like classical shortest paths, it may be computed efficiently using a Dijkstra-type algorithm \cite{Mckenzie2019power}.

The UPD has been proved to be extremely robust to noise, and enjoys excellent performance guarantees for distinguishing between points in the same and different clusters \cite{Little2020}.  These desirable properties suggest its use as a metric in clustering algorithms, such as spectral clustering.

\subsection{Background on Spectral Clustering}
\label{subsec:SpectralClusteringBackground}

For data $X=\{x_{i}\}_{i=1}^{n}\subset\mathbb{R}^{D}$ and metric $\rho:\mathbb{R}^{D}\times\mathbb{R}^{D}\rightarrow [0,\infty)$, define a weighted, undirected graph $\mathcal{G}$ with nodes $X$ and weighted edges $W_{ij}=\exp(-\rho(x_{i},x_{j})^{2}/\sigma^{2})$.  The scaling parameter $\sigma$ can be tuned manually, or set automatically \cite{Zelnik2005_Self}.  Intuitively, there will be a strong edge between $x_{i}, x_{j}$ in $\mathcal{G}$ if and only if $\rho(x_{i}, x_{j})$ is small.

A natural approach to clustering is to partition $\mathcal{G}$ into $K$ communities that are simultaneously ``large" and also pairwise weakly connected.  This may be formulated precisely in terms of the \emph{normalized cuts functional}, which leads to an NP-hard computational problem \cite{Shi2000normalized}.  This graph cut problem may be relaxed by considering eigenvectors of the \emph{graph Laplacian}, which allows to determine natural clusters in $\mathcal{G}$ in polynomial time \cite{Shi2000normalized, Ng2001}.  Indeed, let $D$ be the diagonal degree matrix for $\mathcal{G}$: $D_{ii}=\sum_{j=1}^{n}W_{ij}$.  Let $L=I-D^{-1/2}WD^{-1/2}$ be the (symmetric normalized) graph Laplacian.  Note $L\in\mathbb{R}^{n\times n}$ is positive semi-definite, with a number of zero eigenvalues equal to the number of connected components in $\mathcal{G}$. 

The \emph{spectral clustering algorithm} (Algorithm \ref{alg:SC}) consists in computing the lowest frequency eigenvectors of $L$ (those with smallest eigenvalues) and using them as features in $K$-means clustering \cite{Ng2001}.  Note that we formulate Algorithm \ref{alg:SC} as taking only $W$ and a number of clusters $K$ as an input.  In practice, $W$ must be computed using some metric $\rho$.  While $\rho(x_{i},x_{j})=\|x_{i}-x_{j}\|_{2}$ is common, other metrics may provide superior performance \cite{Elhamifar2011_Sparse, Arias2017_Spectral, Little2020}.

\begin{algorithm}[htb!]
	\caption{\label{alg:SC}Spectral Clustering (SC)}
	\textbf{Input:} $W$, $K$;\\
	\textbf{Output:} $\{y_{i}\}_{i=1}^{n}$
	\begin{algorithmic}[1]
		\STATE Compute the diagonal degree matrix $D\in\mathbb{R}^{n\times n}$.
		\STATE Compute $L=I-D^{-\frac{1}{2}}WD^{-\frac{1}{2}}$.
		\STATE Compute the eigendecomposition  $\{(\phi_{k},\lambda_{k})\}_{k=1}^{n}$, sorted so that $0=\lambda_{1}\le\lambda_{2}\le\dots\le\lambda_{n}$.
		\STATE For $1\leq i\leq n$, let $\tilde{\phi}(x_{i}) = \frac{(\phi_{1}(x_{i}),\phi_{2}(x_{i}),\dots,\phi_{K}(x_{i}))}{||(\phi_{1}(x_{i}),\phi_{2}(x_{i}),\dots,\phi_{K}(x_{i}))||_2}.$
		\STATE Compute labels $\{y_{i}\}_{i=1}^{n}$ by running $K$-means on the data $\{\tilde{\phi}(x_{i}) \}_{i=1}^{n}$ using $K$ as the number of clusters. 
			\end{algorithmic}
\end{algorithm}

\section{Algorithm}

The proposed SRUSC method (Algorithm \ref{alg:SR_USC}) consists in performing spectral clustering using the UPD $\rho_{\infty}$ and a spatially regularized graph Laplacian.  More precisely, let
\begin{align}\label{eqn:W_UPC_spatial}W_{ij}=
\begin{cases}
\exp(-\rho_{\infty}(x_{i},x_{j})^{2}/\sigma^{2}), &x_{i}\in B_{r}(x_{j}),\\
0, &x_{i}\notin B_{r}(x_{j}),
\end{cases}
\end{align}where $B_{r}$ is the set of all pixels whose spatial coordinates lie inside the square with side lengths $r$ centered at $x_{i}$.  The constraint that $W_{ij}=0$ if $x_{i}\notin B_{r}(x_{j})$ or $x_{j}\notin B_{r}(x_{i})$ enforces spatial regularity in the graph: a pixel can only be connected to spatially proximal pixels.  Spatial regularization has been shown to improve clustering performance for HSI, by producing clusters that are smoother with respect to the underlying spatial structure \cite{Cahill2014_Schroedinger, Murphy2020_Spectral}. It also has a sparsifying effect, since $W$ has only $r^{2}n$ non-zero entries.  This affords substantial computational advantage in the subsequent eigenvector calculation when $r^{2}\ll n$.

\begin{algorithm}[htb!]
	\caption{\label{alg:SR_USC}Spatially Regularized Ultrametric SC (SRUSC)}
	\textbf{Input:} $\{x_{i}\}_{i=1}^{n},\sigma, K$; \\
	\textbf{Output:} $\{y_{i}\}_{i=1}^{n}$
		\begin{algorithmic}[1]
	    \STATE Construct $W$ as in (\ref{eqn:W_UPC_spatial}).
        \STATE Run Algorithm \ref{alg:SC} with inputs $W$, $K$.
		\end{algorithmic}
\end{algorithm}

\subsection{Computational Complexity}
\label{subsec:CC}

Note that the graph Laplacian $L$ constructed in the proposed method is $r^{2}$-sparse.  Since the cost of computing $\rho_{\infty}(x_{i},x_{j})$ is proportional to the number of edges in the underlying graph $\mathcal{G}$, the complexity of computing $L$ is $O(r^{2}nk)$, where $k$ is the number of nearest neighbors in $\mathcal{G}_{0}$.  It is known that $k\sim\log(n)$ is sufficient to ensure that with high probability the UPD on a $k$-nearest neighbors graph and UPD on the fully connected graph are the same \cite{Gonzalez2003_Clustering, Little2020}.  For such a $k$, the computation of $L$ is $O(r^{2}n\log(n))$.  Once $L$ is computed, getting the $K=O(1)$ lowest frequency eigenvectors is $O(r^{2}n)$ using iterative methods.  Finally, running $K$-means via Lloyd's algorithm on these eigenvectors is $O(n)$ if the number of iterations is constant with respect to $n$.  Thus, the overall algorithm has complexity $O(r^{2}n\log(n))$.  
\label{sec:Algorithm}

\subsection{Discussion of Parameters}
\label{subsec:Parameters}

The main parameters of the proposed method are $\sigma$, $r$, and $K$.  Many automated methods exist for determining the scaling parameter $\sigma$ in spectral clustering \cite{Zelnik2005_Self}, and spatially regularized spectral graph methods are typically somewhat robust to the choice of spatial radius $r$.  On the other hand, the number of clusters $K$ is notoriously challenging to estimate, particularly for data with nonlinear or elongated shapes.  

A commonly employed heuristic in spectral clustering is the \emph{eigengap} heuristic, which estimates $\hat{K}=\argmax_{k}\lambda_{k+1}-\lambda_{k}$.  However, this depends strongly on $\sigma$.  One can instead consider a \emph{multiscale eigengap} \cite{Little2015_Multiscale, Little2020}, which simultaneously maximizes over the eigenvalue index and over $\sigma$: \[\hat{K}=\argmax_{\sigma\in S}\left(\argmax_{k}\lambda_{k+1}(\sigma)-\lambda_{k}(\sigma)\right),\]where $S=\{\sigma_{\ell}\}_{\ell=1}^{L}$ and $\lambda_{i}(\sigma)$ is the $i^{th}$-largest eigenvalue of $L$ when it is computed using $\sigma$.  We show in Section \ref{subsec:EstimatingK} that this approach is effective when using UPD for HSI.

\section{Experimental Analysis}
\label{sec:ExperimentalAnalysis}

To validate the proposed method, we perform clustering experiments on four data sets: two synthetic HSI, and two real HSI.      

\begin{figure}[!htb]
\centering
\begin{subfigure}{0.23\textwidth}
    \includegraphics[width=\textwidth]{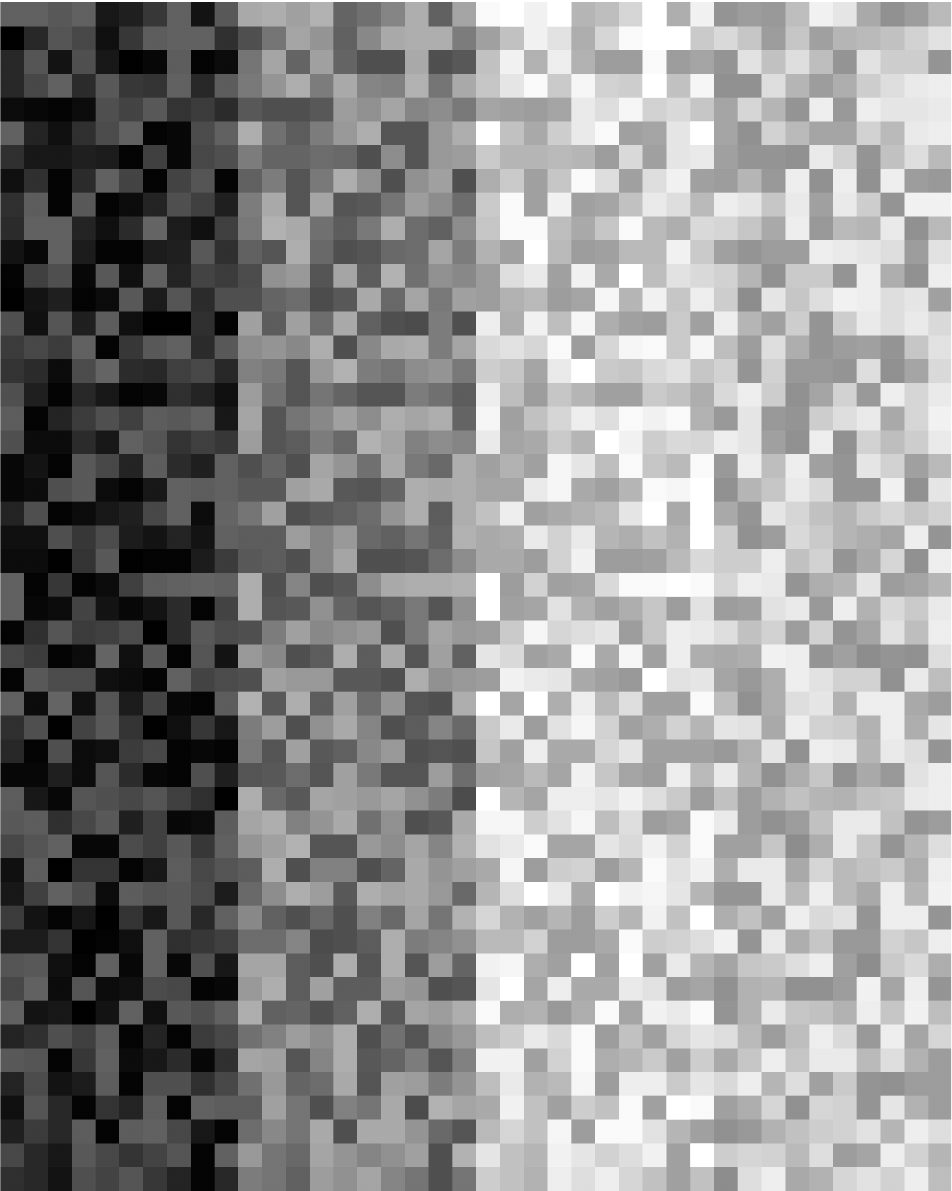}
    \subcaption{Projection on the 1st PC}
\end{subfigure}
\begin{subfigure}{0.23\textwidth}
    \includegraphics[width=\textwidth]{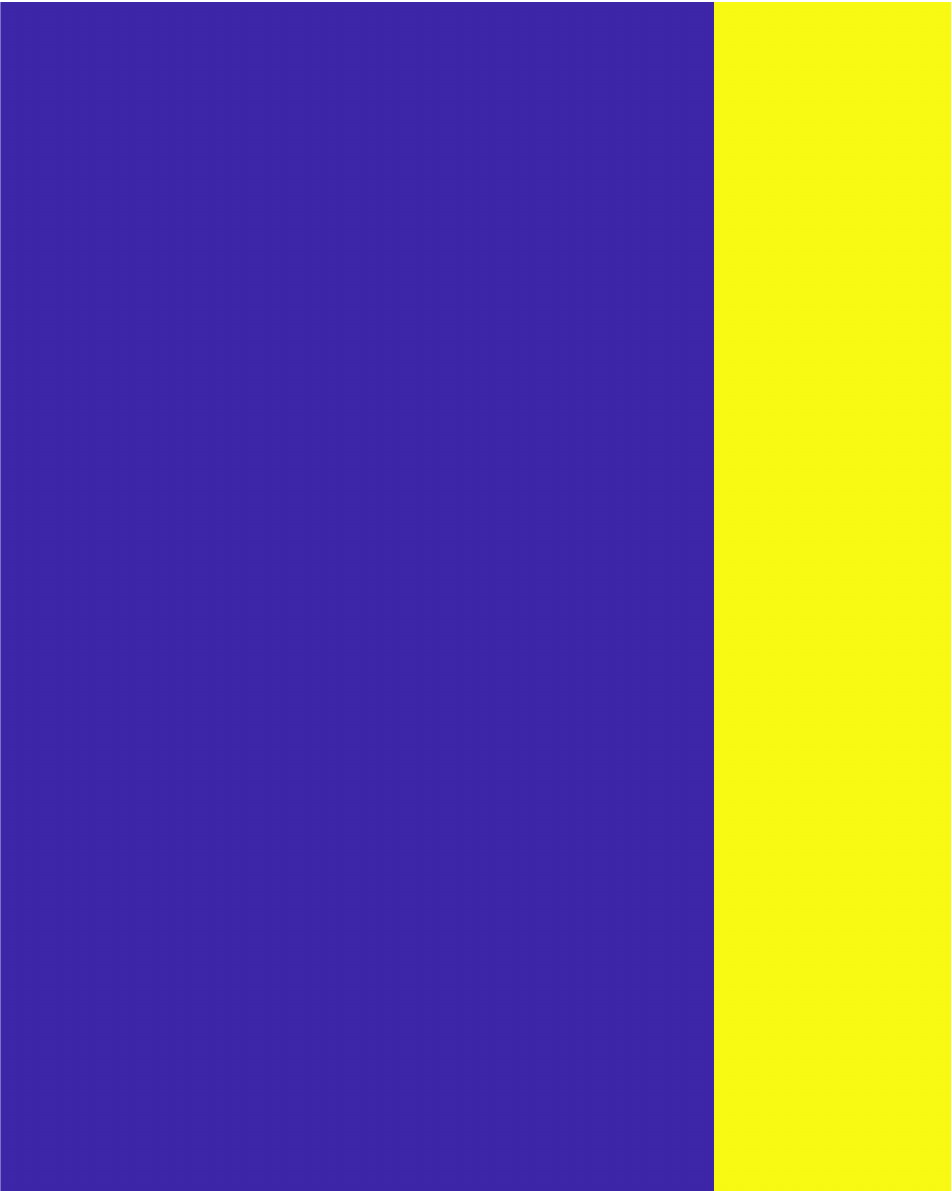}
    \subcaption{GT}
\end{subfigure}
\caption{\label{fig:FS}The four spheres synthetic data is $40\times 50\times 200$ and contains two clusters.  The projection onto the first principal component  is in (a), the ground truth labels in (b).  A spatial radius of $r=30$ was used in SRUSC.}
\end{figure}

The first synthetic data set is generated by uniformly sampling 500 points from 4 2-dimensional spheres of radii $1.7$ and centers $(1,3), (1,5), (1,7)$, and $(5,5)$, respectively, embedded in $\mathbb{R}^{200}$.  There are 2 clusters.  The first consists of the union of the spheres  with centers $(1,3), (1,5)$ and $(1,7)$, while the second cluster consists of the sphere with center $(5,5)$.  Each sphere contains 500 points, spatially arrayed to be $10\times 50$. These clusters are concatenated spatially into a $40\times 50\times 200$ synthetic HSI;  see Fig. \ref{fig:FS}.  Methods based on Euclidean distance are expected to fail because the spectral diameter of the first cluster is large when measured with Euclidean distance. 

\begin{figure}[!htb]
\centering
\begin{subfigure}{0.23\textwidth}
    \includegraphics[width=\textwidth]{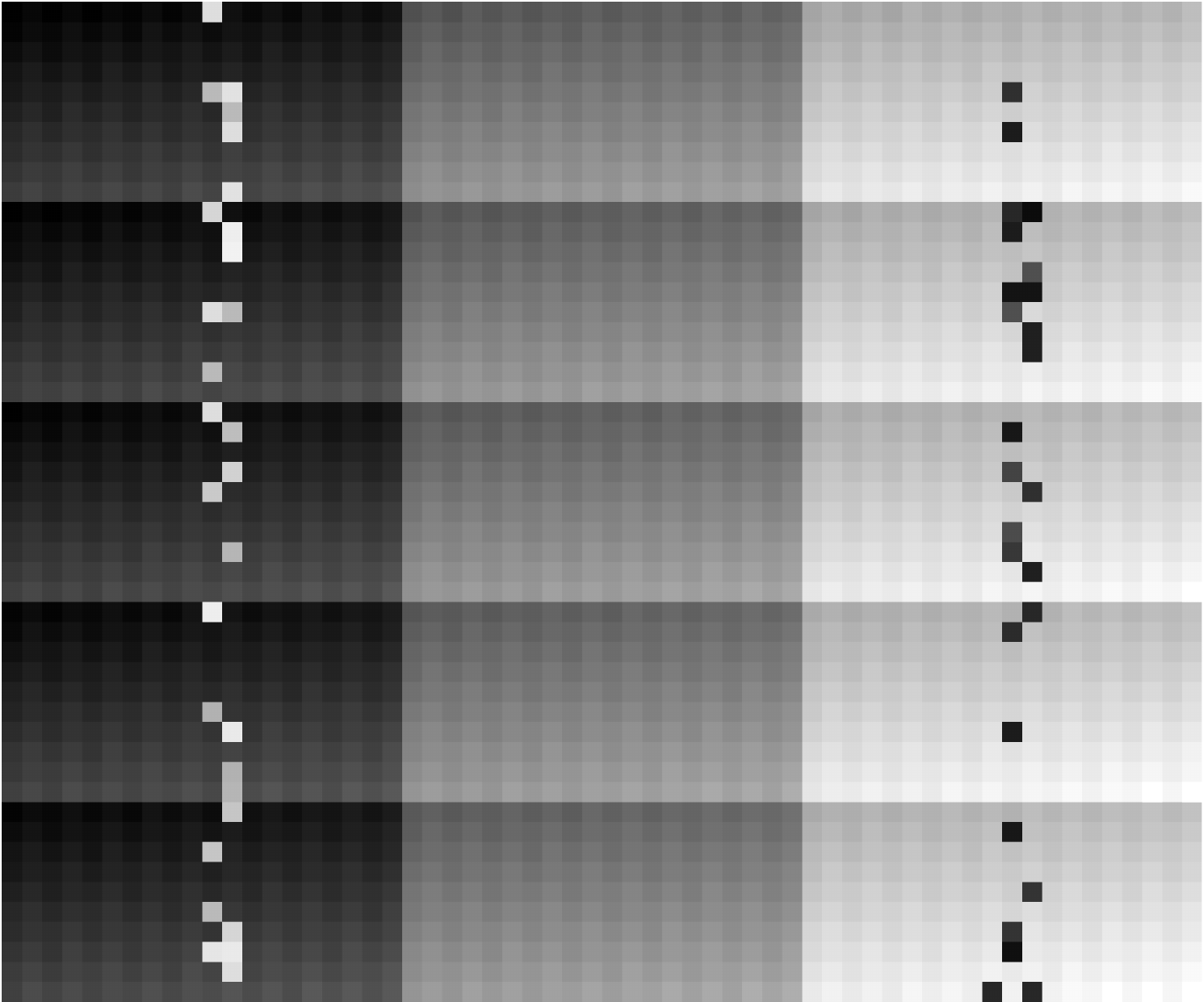}
    \subcaption{Projection on the 1st PC}
\end{subfigure}
\begin{subfigure}{0.23\textwidth}
    \includegraphics[width=\textwidth]{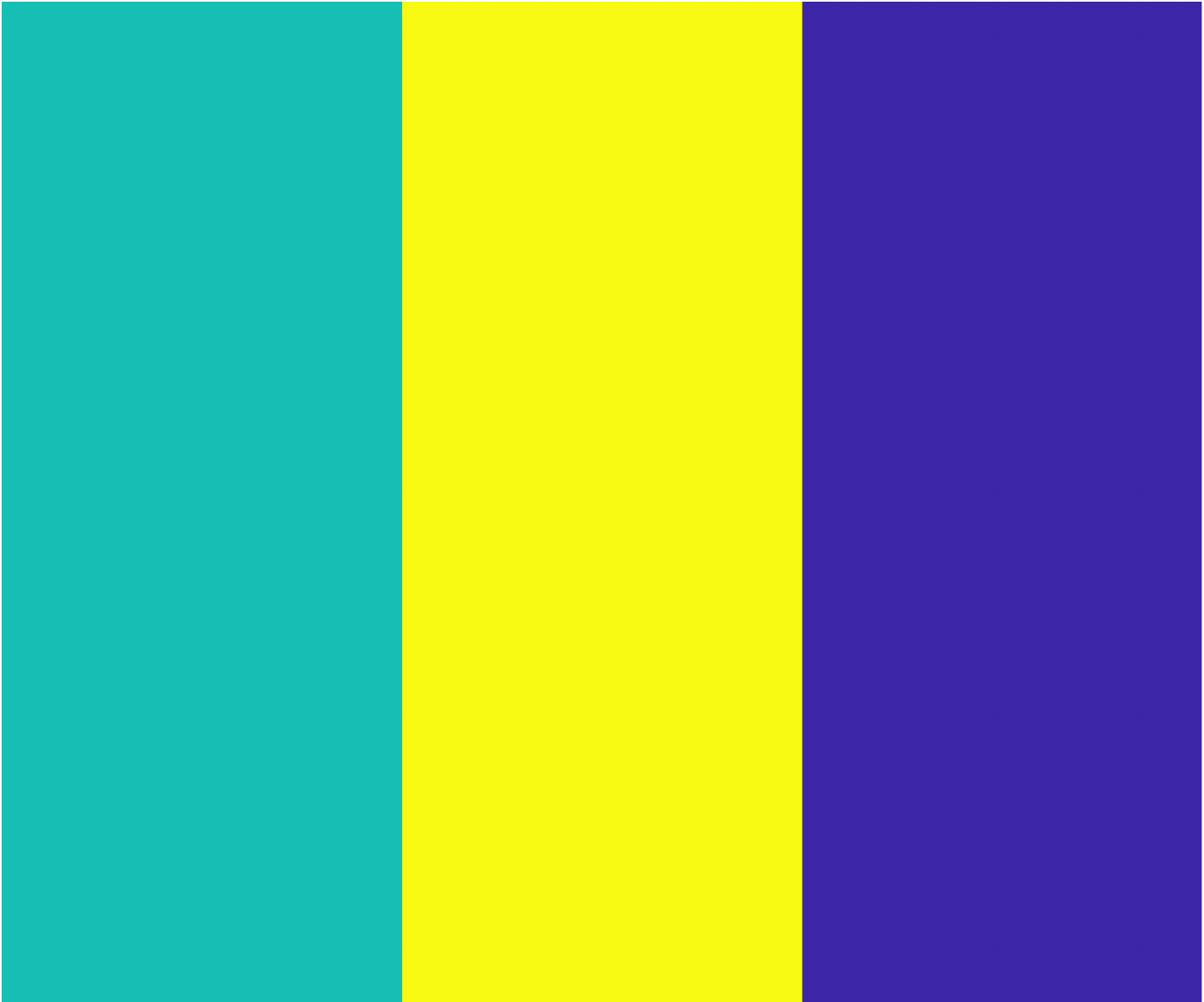}
    \subcaption{GT}
\end{subfigure}
\caption{\label{fig:TC}The three cubes synthetic data is $60\times 50\times 200$ and contains 3 clusters.  The projection onto the first principal component  is in (a), the ground truth labels in (b).   A spatial radius of $r=30$ was used in SRUSC.}
\end{figure}

The second synthetic data set is generated by randomly rotating three copies of a uniform grid sample of $[0,1]^{3}$ into $\mathbb{R}^{199}$, then separating them in the $200^{\text{th}}$ dimension by translating the second cube by $(0,0,\dots,0,.1)$ and the third cube by $(0,0,\dots,0,.2)$. Each cube contains 1000 points, spatially arrayed to be $20\times 50$. These clusters are concatenated spatially into a $60\times 50\times 200$ synthetic HSI; see Fig. \ref{fig:TC}.  To demonstrate the necessity of spatial regularization, we randomly select 30 points from the middles of cube 1 and cube 3 and swap them.  This can be understood as a kind of noise, to which we expect spatially regularized methods to be robust.

\begin{figure}[!htb]
\centering
\begin{subfigure}{0.23\textwidth}
    \includegraphics[width=\textwidth]{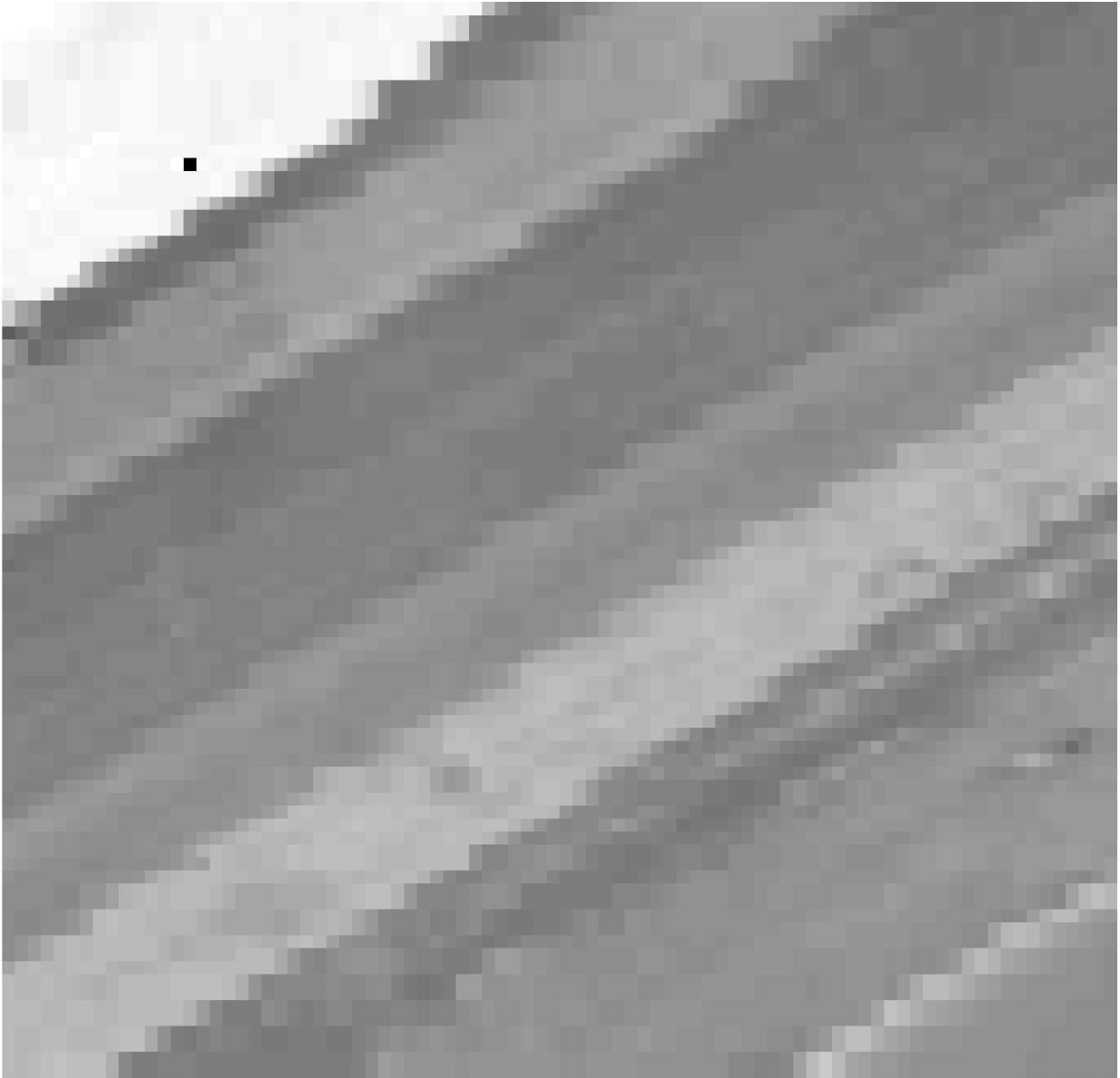}
    \subcaption{Projection on the 1st PC}
\end{subfigure}
\begin{subfigure}{0.23\textwidth}
        \includegraphics[width=\textwidth]{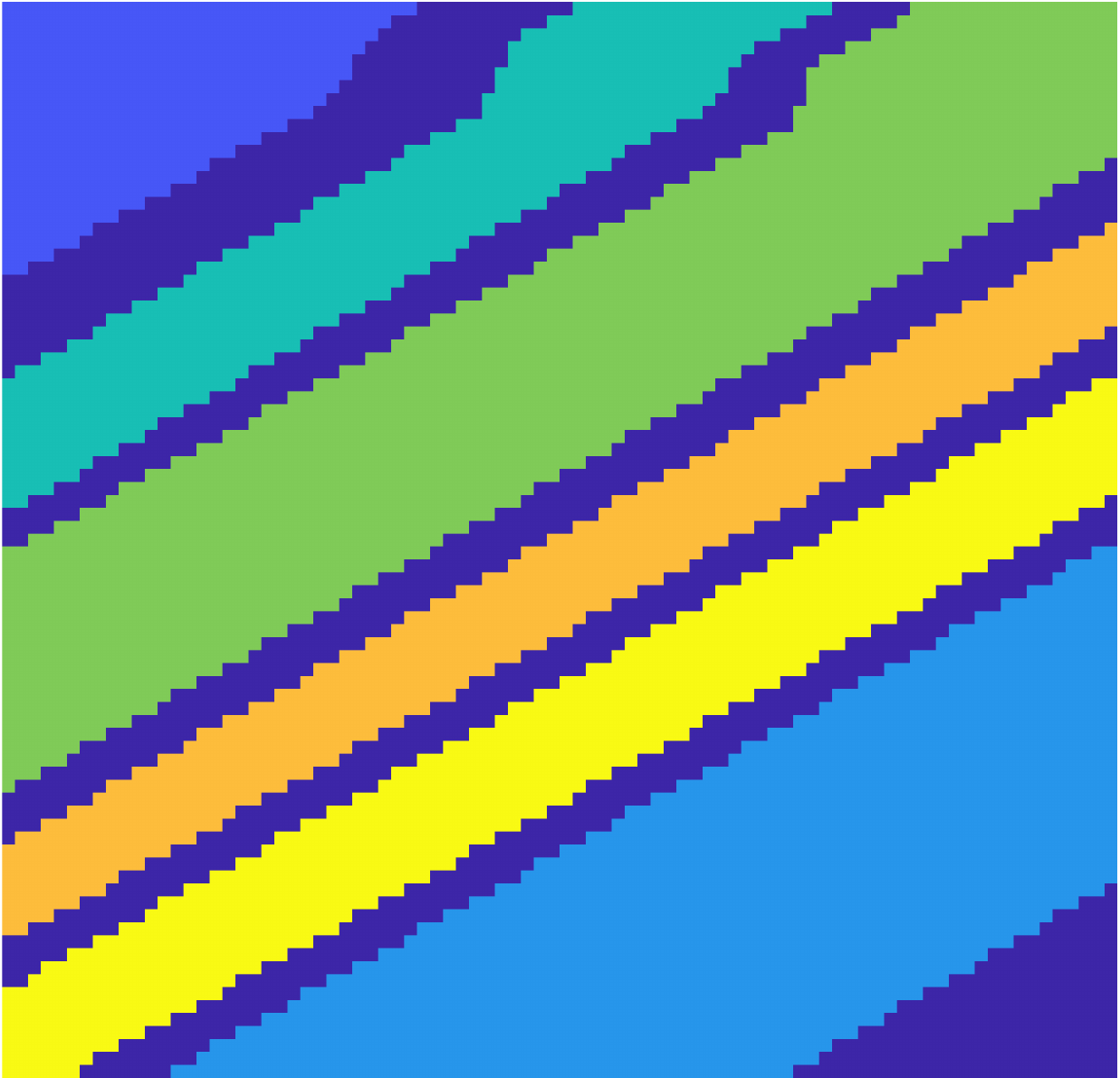}
    \subcaption{Salinas A Ground Truth}
\end{subfigure}
\caption{\label{fig:SalinasA}  The Salinas A data set is a $83\times 86$ real HSI taken by the 224-band AVIRIS sensor over Salinas Valley, California.  There are 6 clusters in the ground truth.  The projection onto the first principal component  is in (a), the ground truth labels in (b).   A spatial radius of $r=65$ was used in SRUSC.}
\end{figure}

\begin{figure}[!htb]
\centering
\begin{subfigure}{0.23\textwidth}
    \includegraphics[width=\textwidth]{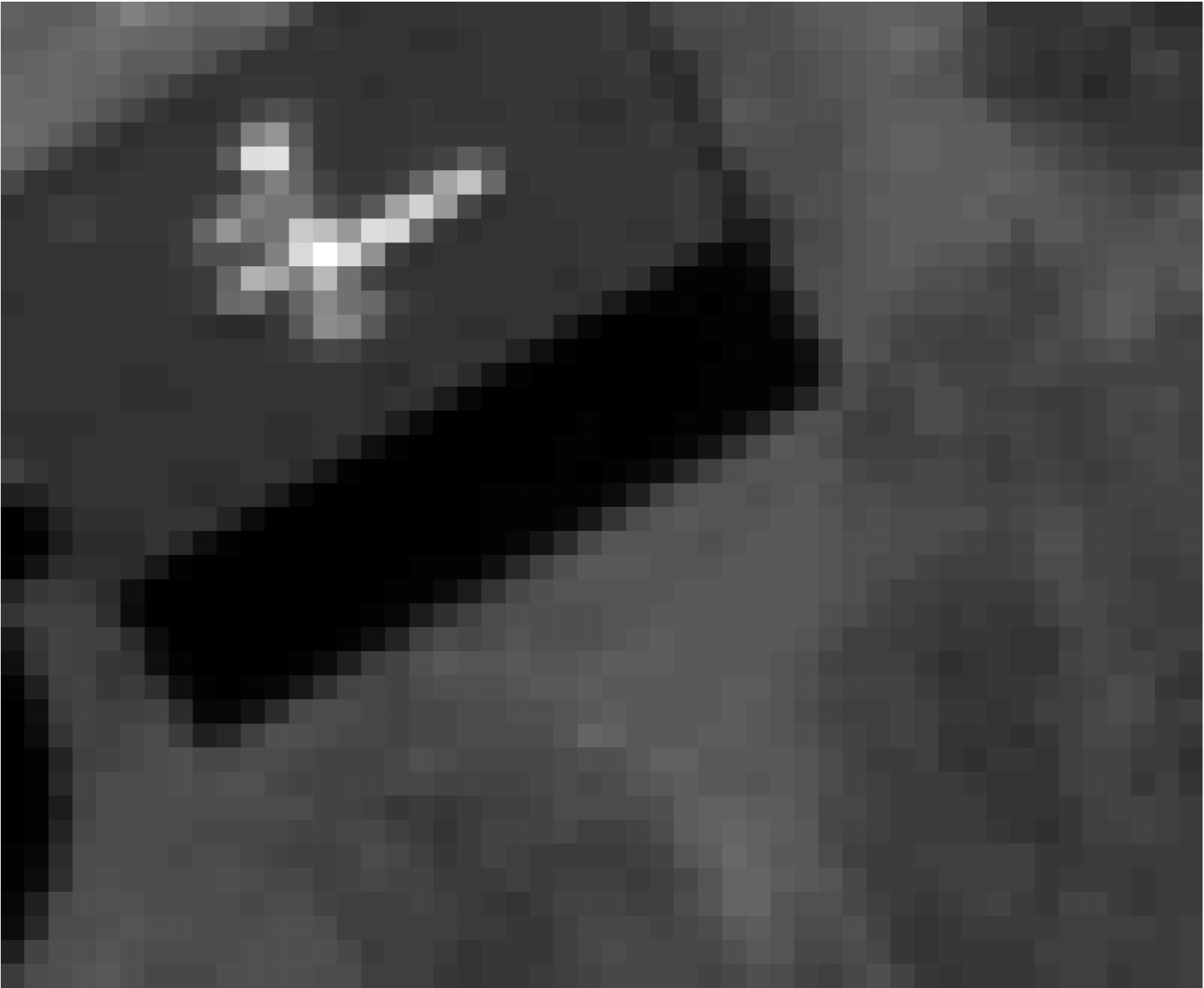}
    \subcaption{Projection on the 1st PC}
\end{subfigure}
\begin{subfigure}{0.23\textwidth}
        \includegraphics[width=\textwidth]{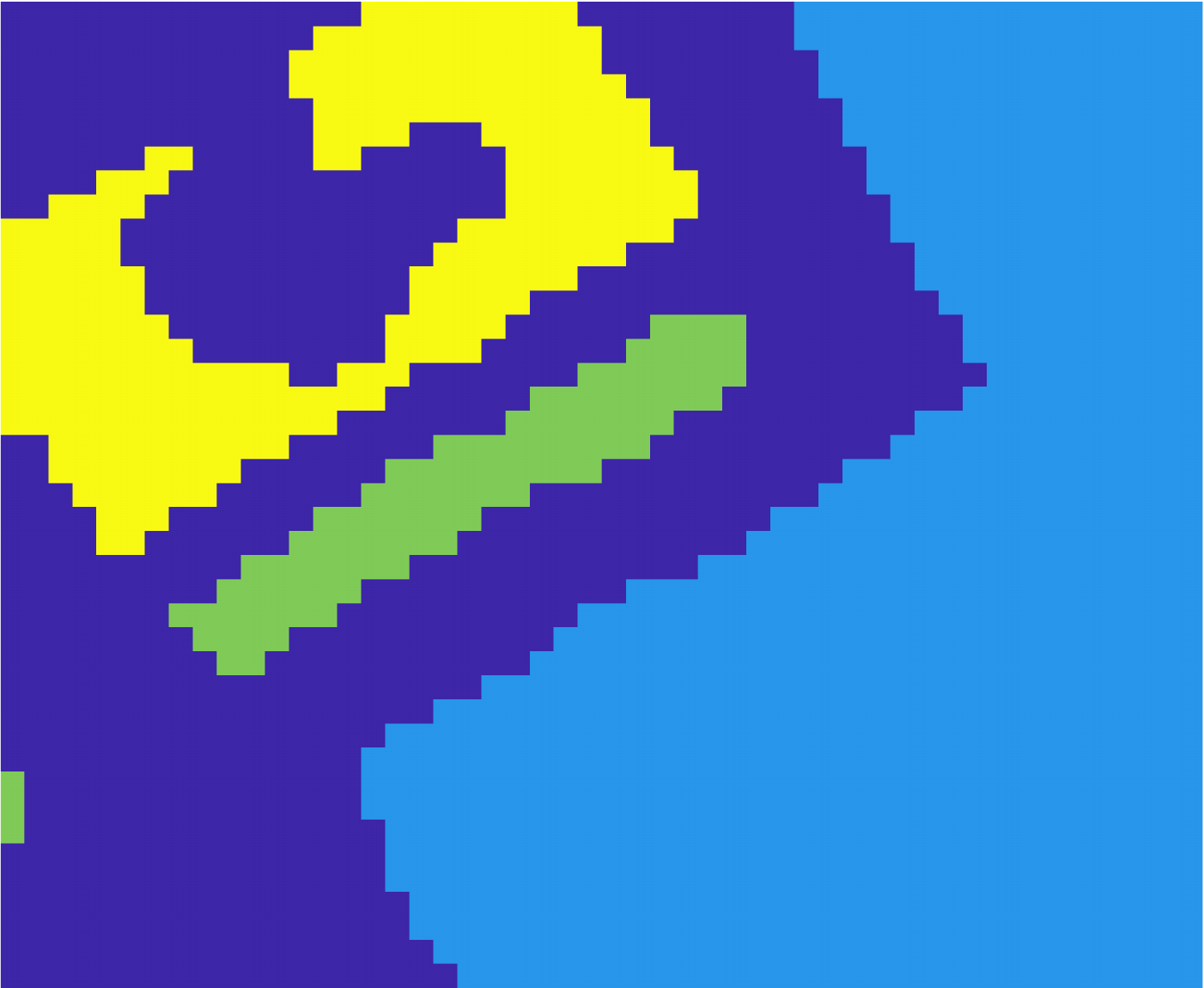}
    \subcaption{Pavia U Ground Truth}
\end{subfigure}
\caption{\label{fig:PaviaU}  The Pavia U data set is a $40\times 51$ subset of the full HSI acquired by the 103-band ROSIS sensor over Pavia University.  There are 3 clusters in the ground truth.  The projection onto the first principal component  is in (a), the ground truth labels in (b).  A spatial radius of $r=30$ was used in SRUSC.}
\end{figure}

We also consider two real HSI data sets: the Salinas A and Pavia U data sets\footnote{\url{http://www.ehu.eus/ccwintco/index.php/Hyperspectral_Remote_Sensing_Scenes}}.  Visualizations of the real HSI, along with their partial ground truth, are in Fig. \ref{fig:SalinasA}, \ref{fig:PaviaU}, respectively.  Note that both of these data sets contain a relatively small number of classes; in the case of Pavia U we chose to crop a small subregion to use for experiments, due to both the well-documented challenges of using too many ground truth classes for unsupervised HSI clustering \cite{Zhu2017_Unsupervised}, and in order to ensure that most pixels in the image had ground truth labels.

\subsection{Comparison Methods}

We compare with four benchmark clustering methods, as well as four state-of-the-art methods.  The benchmark methods are  \emph{$K$-means} clustering \cite{Bishop2006}; \emph{$K$-means} clustering on data that has been dimension-reduced with \emph{PCA}, using the top $K$ principal components; \emph{Gaussian mixture models} \cite{Bishop2006}; and \emph{spectral clustering} using Euclidean distance \cite{Ng2001}.  The state-of-the-art methods are diffusion learning \cite{Murphy2019_Unsupervised, Maggioni2019_Learning}; density peaks clustering \cite{Rodriguez2014}; nonnegative matrix factorization \cite{Gillis2015}; and local covariance matrix representation \cite{Fang2018}.  For all experiments, we assume the number of clusters $K$ is known a priori; in Section \ref{subsec:EstimatingK}, we show how the proposed method can estimate $K$.

\subsection{Clustering Accuracy}

\begin{figure}[!htb]
    \centering
     \begin{subfigure}{0.09\textwidth}
        \includegraphics[width=\textwidth]{images/FC/FC_gt-crop.pdf}
        \subcaption{GT}
    \end{subfigure}
    \begin{subfigure}{0.09\textwidth}
        \includegraphics[width=\textwidth]{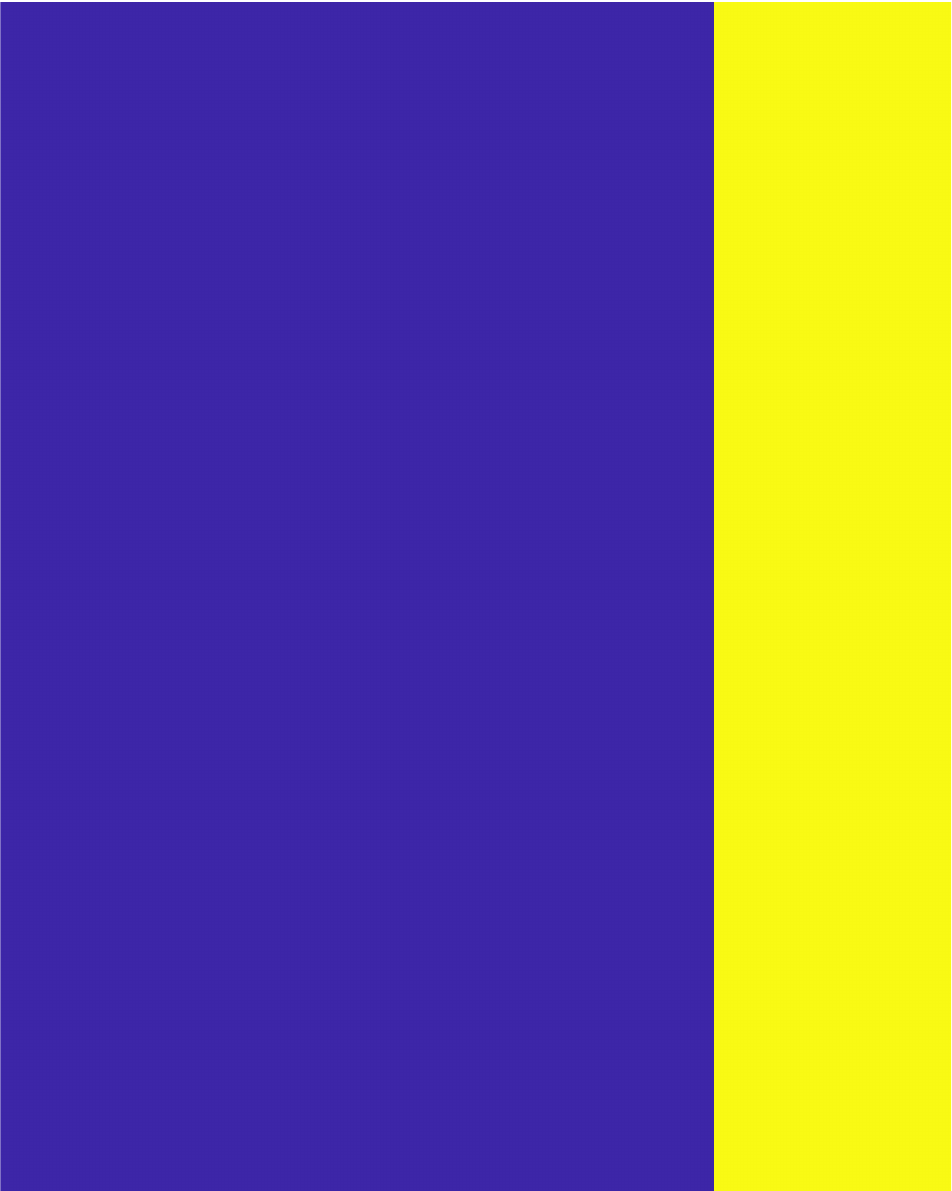}
        \subcaption{KM}
    \end{subfigure}
    \begin{subfigure}{0.09\textwidth}
        \includegraphics[width=\textwidth]{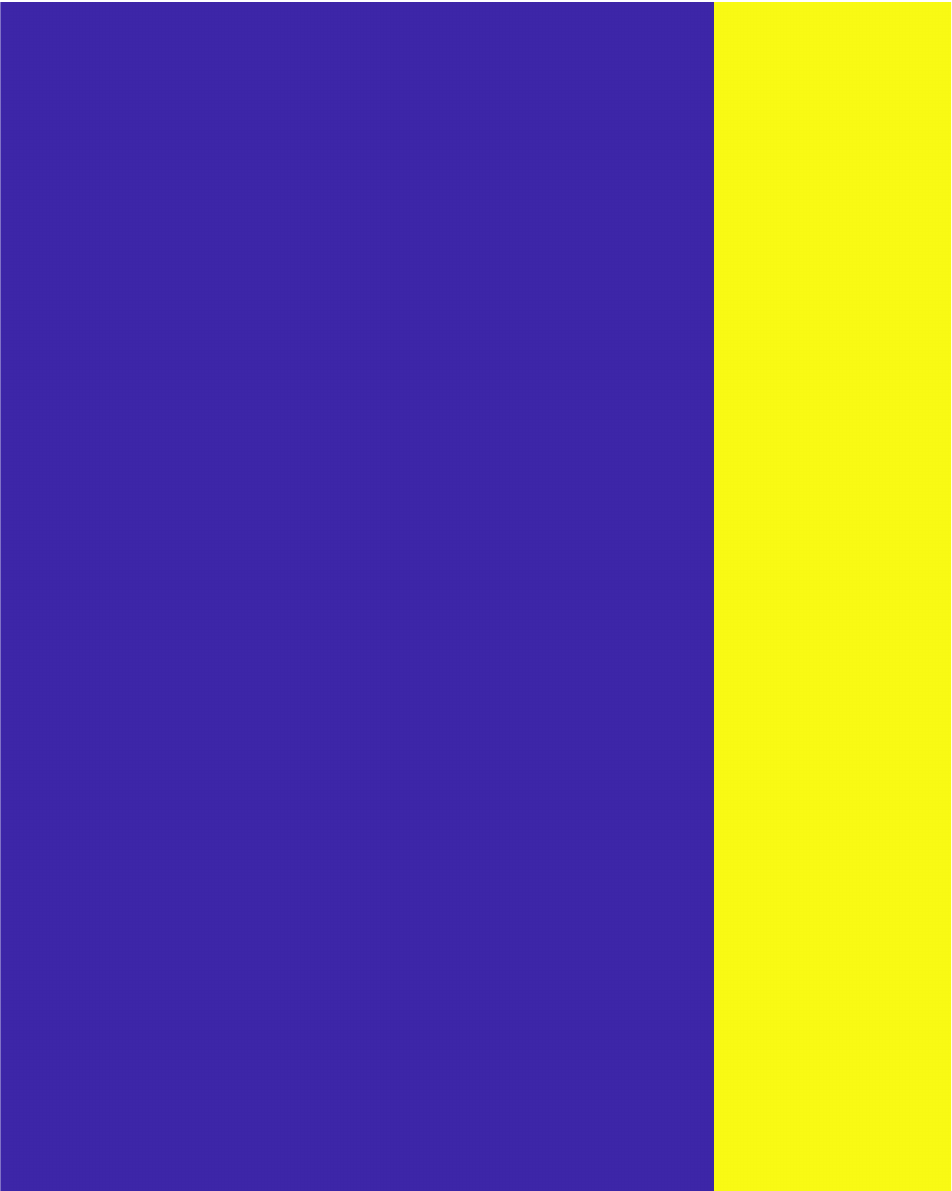}
        \subcaption{PCA}
    \end{subfigure}
    \begin{subfigure}{0.09\textwidth}
        \includegraphics[width=\textwidth]{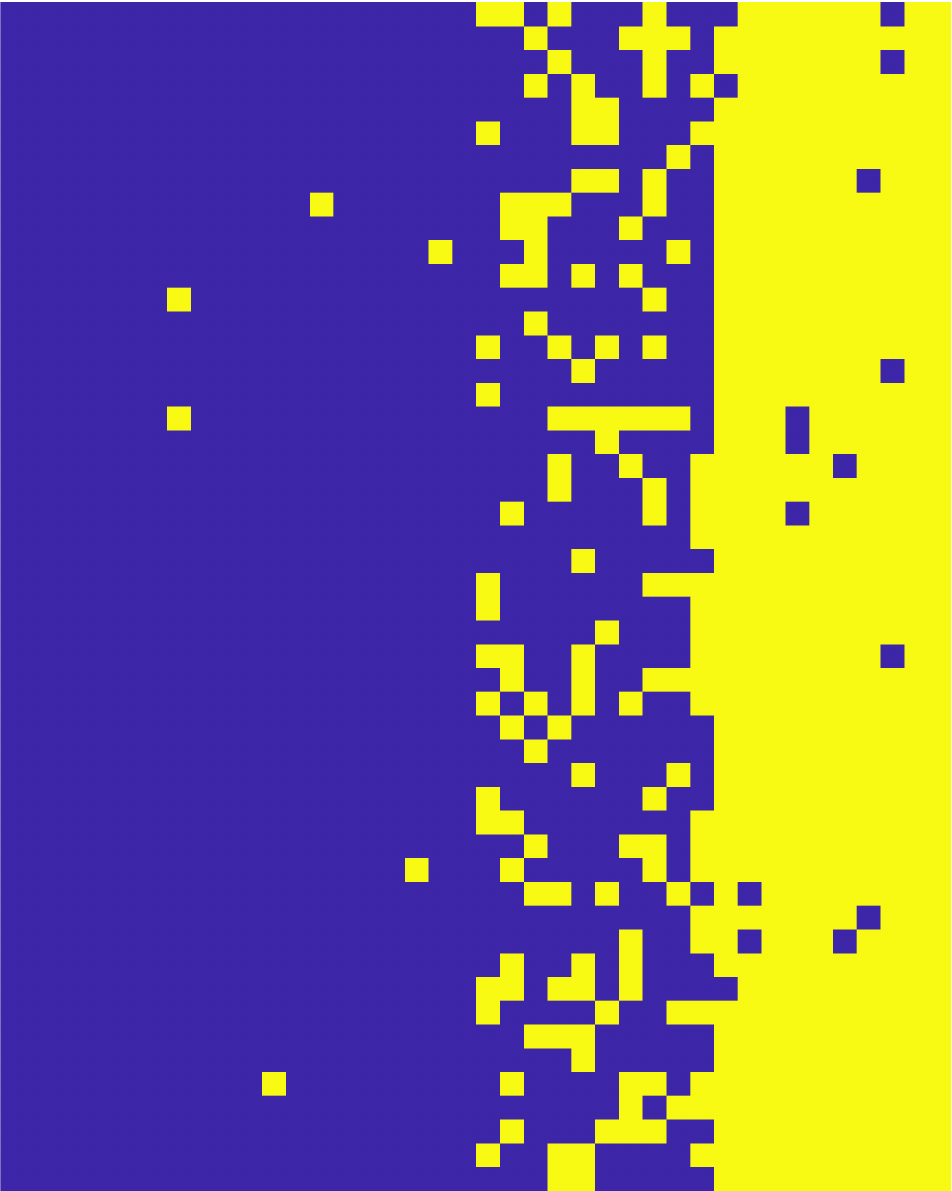}
        \subcaption{GMM}
    \end{subfigure}
     \begin{subfigure}{0.09\textwidth}
        \includegraphics[width=\textwidth]{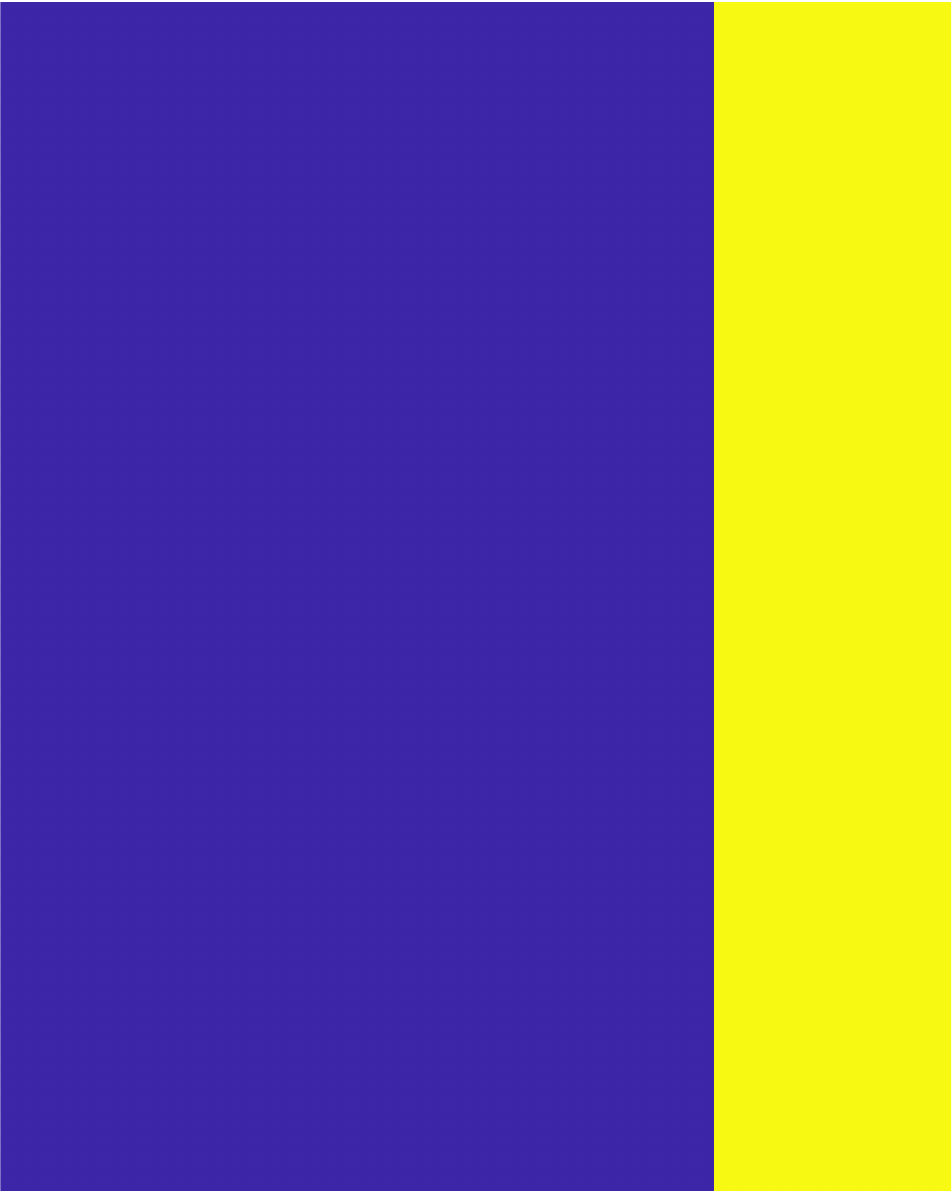}
        \subcaption{SC}
    \end{subfigure}
     \begin{subfigure}{0.09\textwidth}
        \includegraphics[width=\textwidth]{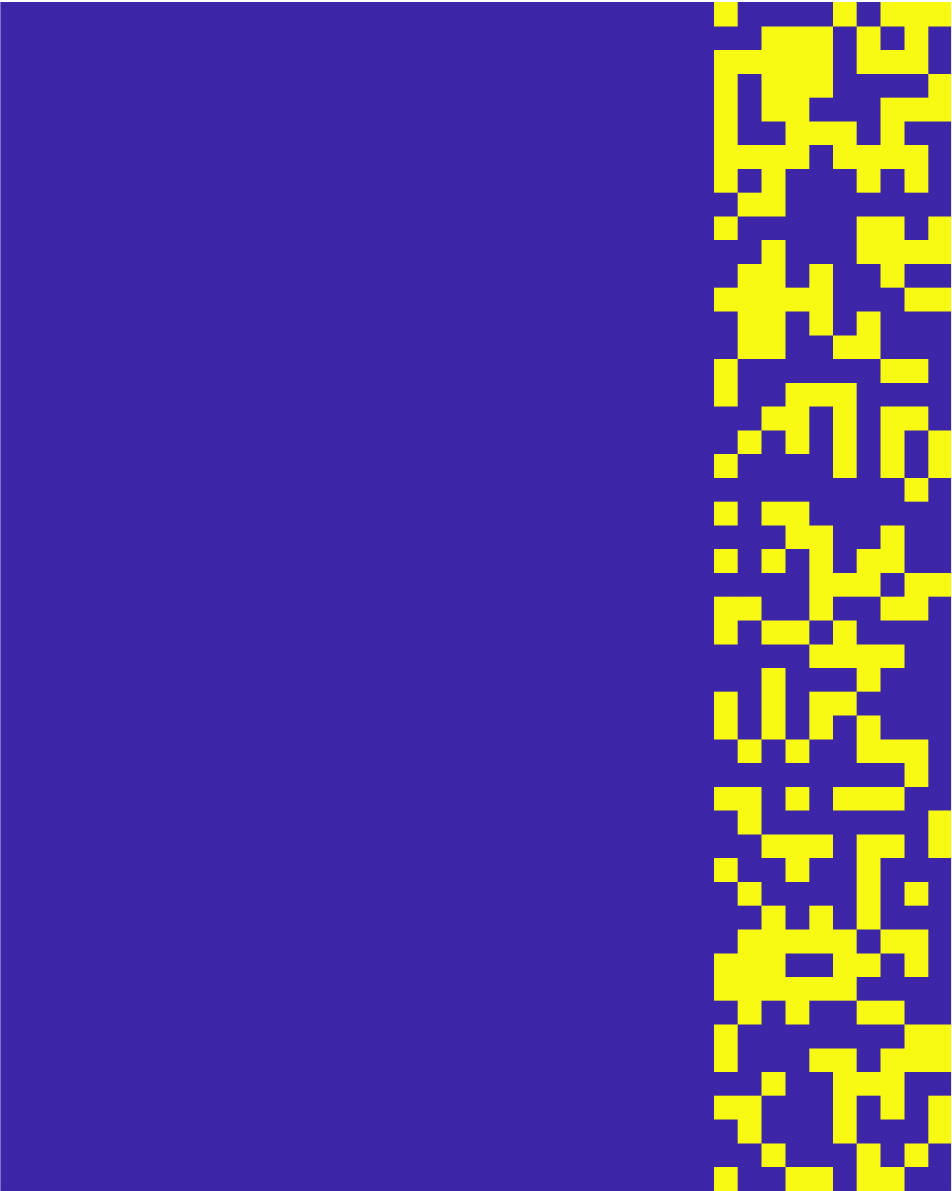}
        \subcaption{DL}
    \end{subfigure}
     \begin{subfigure}{0.09\textwidth}
        \includegraphics[width=\textwidth]{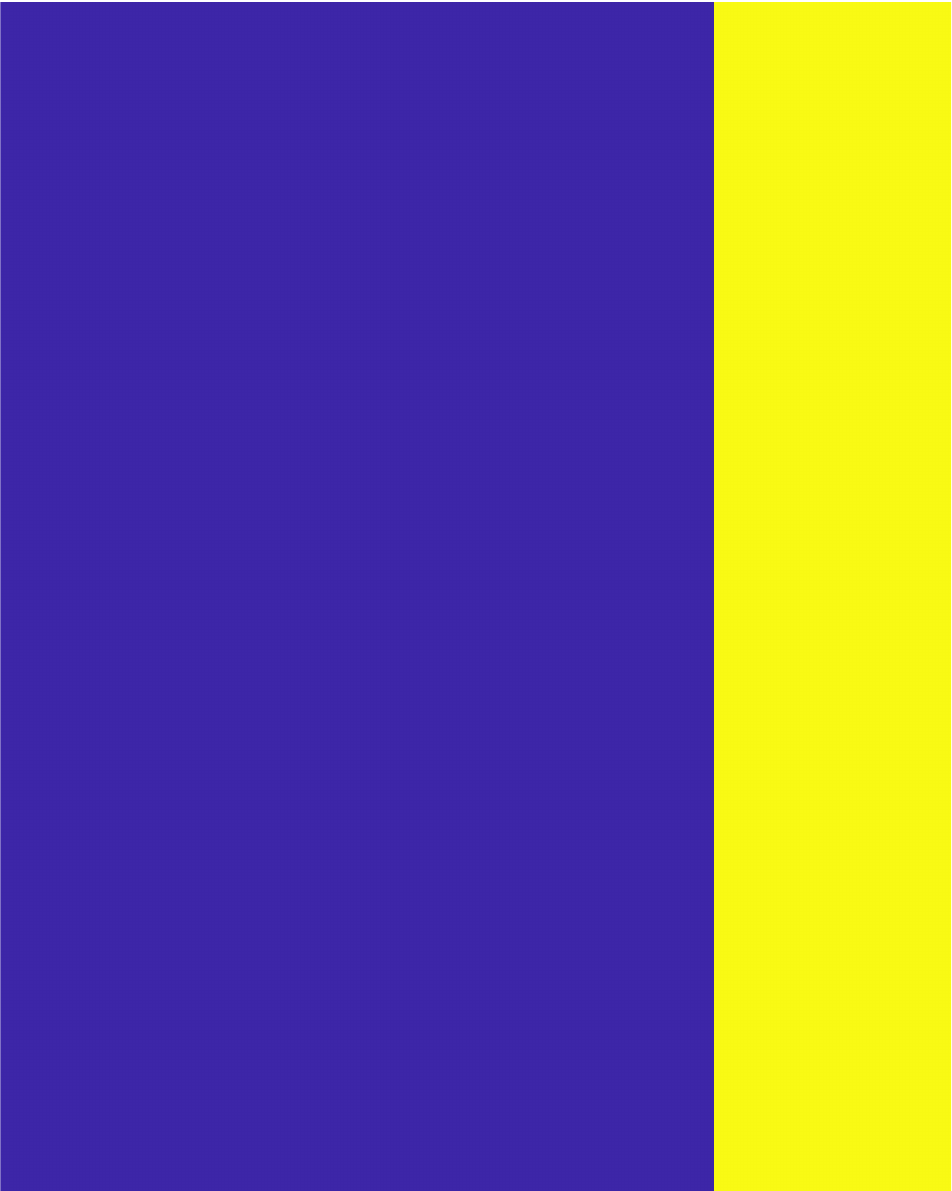}
        \subcaption{FSFDPC}
    \end{subfigure}
     \begin{subfigure}{0.09\textwidth}
        \includegraphics[width=\textwidth]{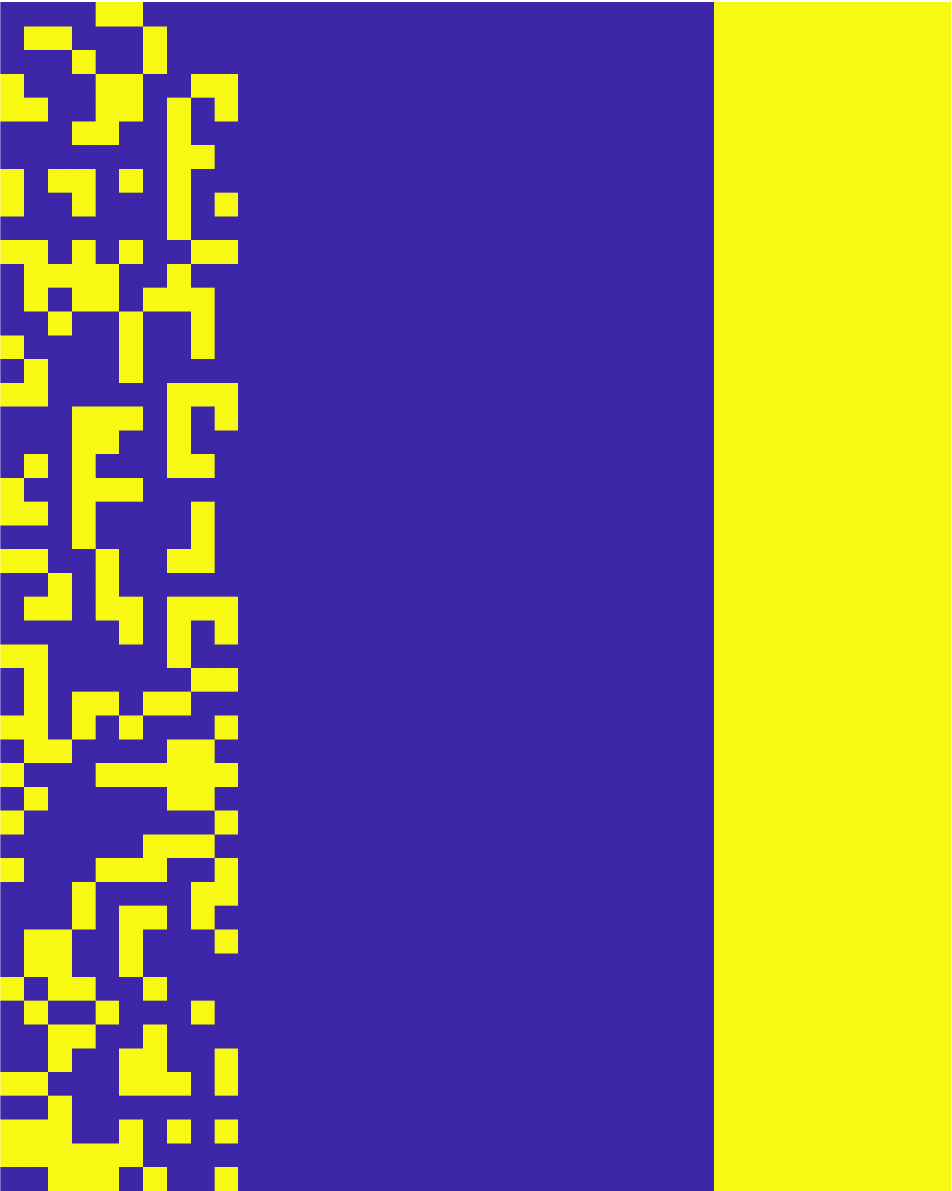}
        \subcaption{NMF}
    \end{subfigure}
     \begin{subfigure}{0.09\textwidth}
        \includegraphics[width=\textwidth]{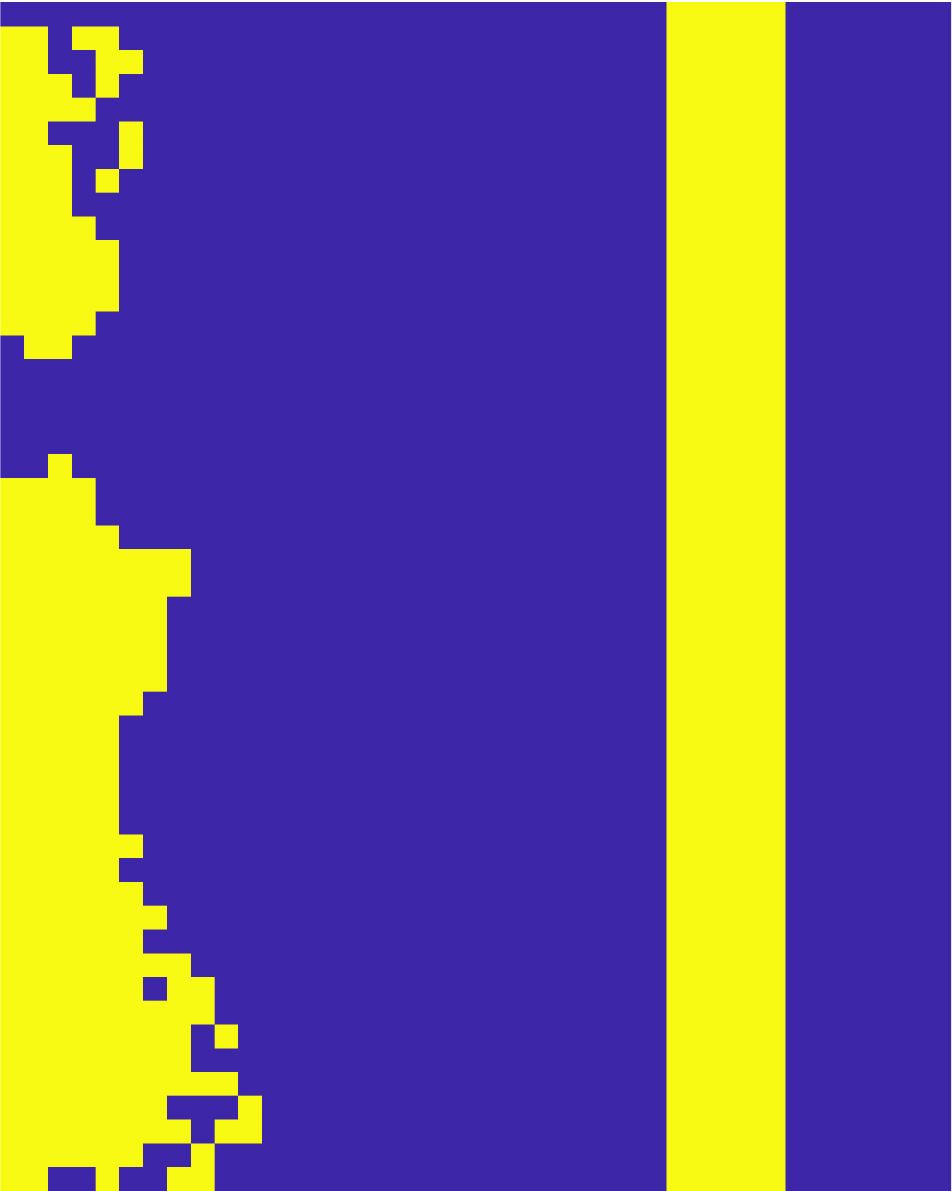}
        \subcaption{LCMR}
    \end{subfigure}
     \begin{subfigure}{0.09\textwidth}
        \includegraphics[width=\textwidth]{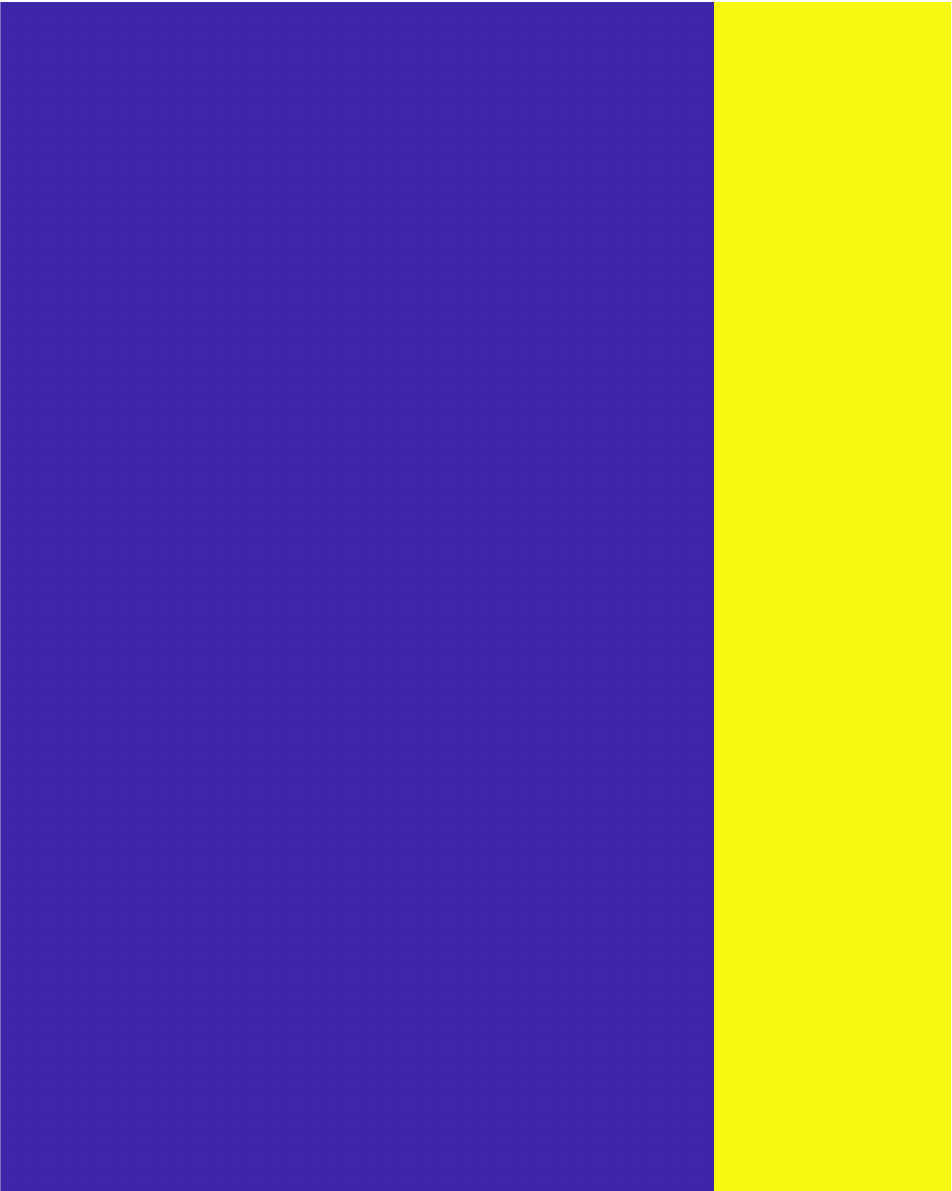}
        \subcaption{SRUSC}
    \end{subfigure}
    \caption{\label{fig:FourCircles_Results}  For the four spheres synthetic data set, several methods achieve perfect accuracy, including the proposed method.}
    \end{figure}

\begin{figure}[!htb]
    \centering
     \begin{subfigure}{0.09\textwidth}
        \includegraphics[width=\textwidth]{images/TC/TC_gt-crop.pdf}
        \subcaption{GT}
    \end{subfigure}
    \begin{subfigure}{0.09\textwidth}
        \includegraphics[width=\textwidth]{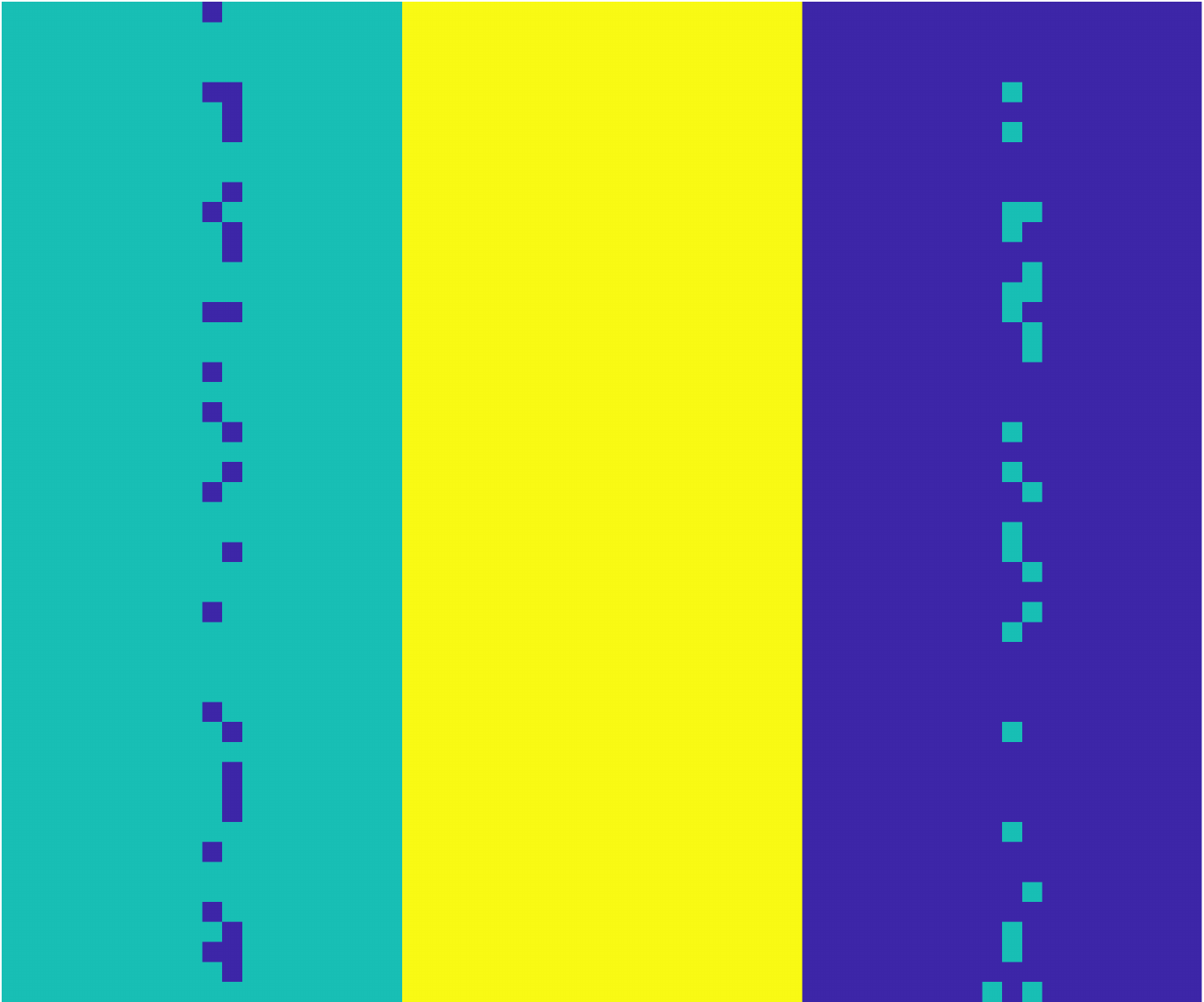}
        \subcaption{KM}
    \end{subfigure}
    \begin{subfigure}{0.09\textwidth}
        \includegraphics[width=\textwidth]{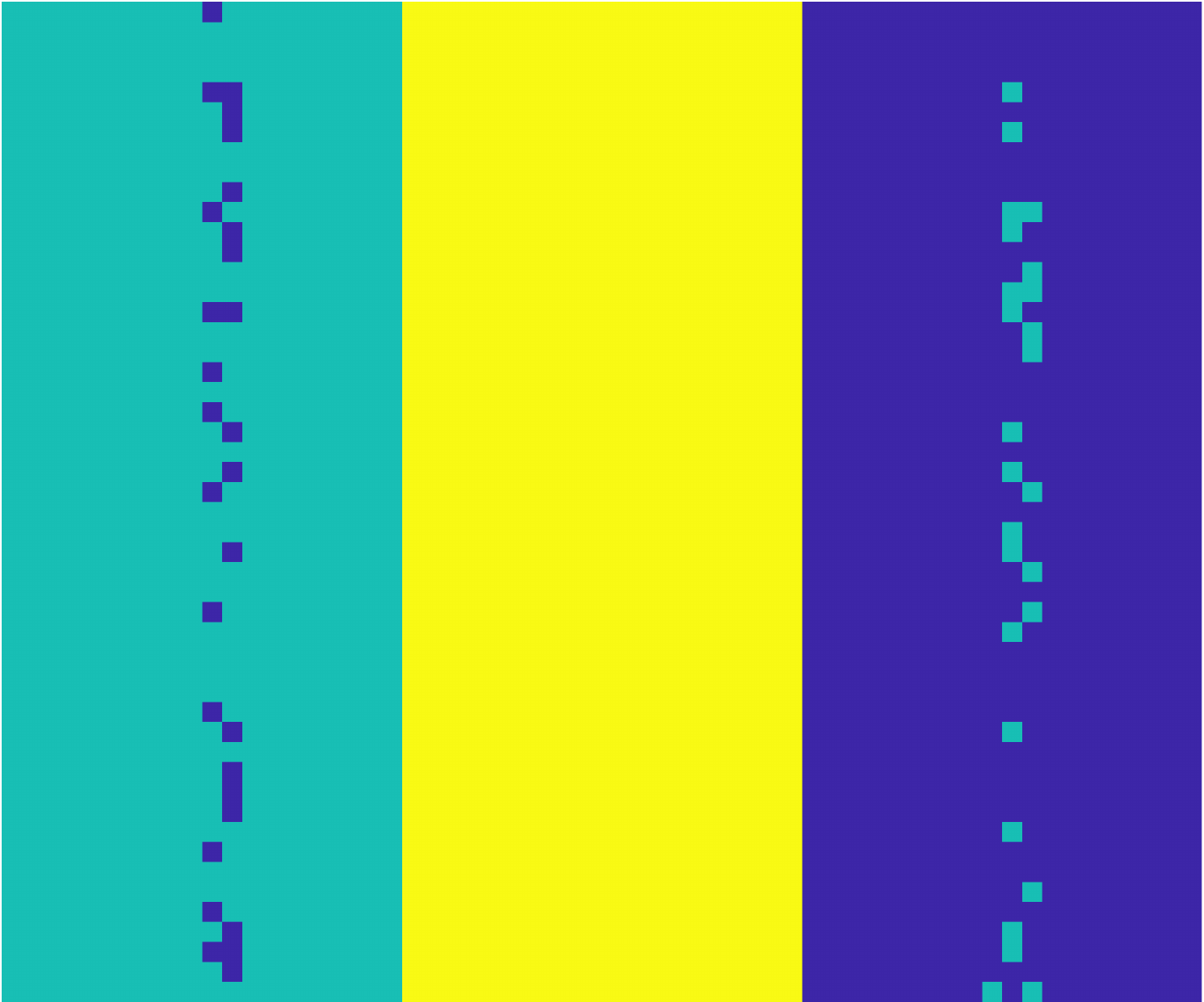}
        \subcaption{PCA}
    \end{subfigure}
    \begin{subfigure}{0.09\textwidth}
        \includegraphics[width=\textwidth]{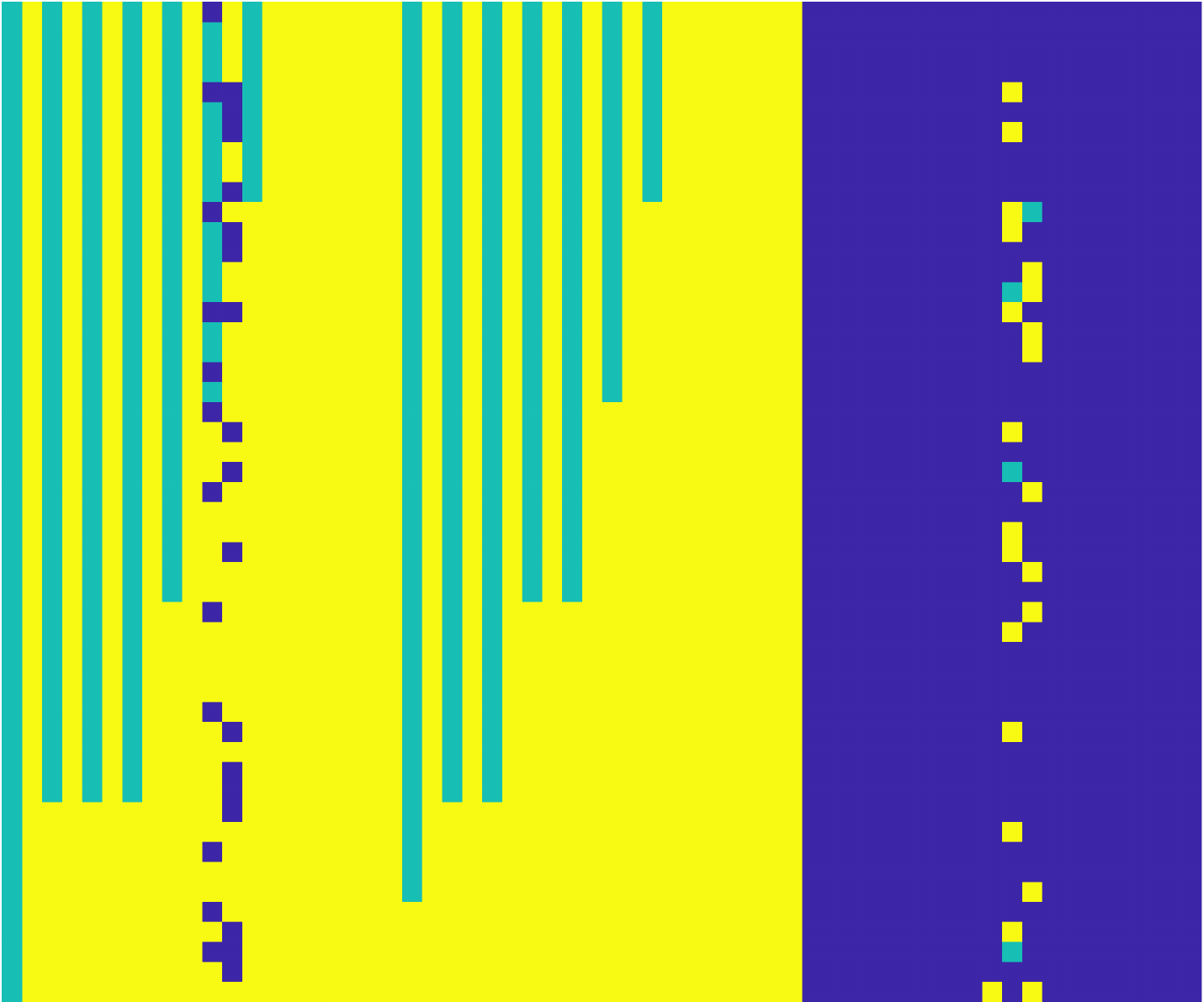}
        \subcaption{GMM}
    \end{subfigure}
     \begin{subfigure}{0.09\textwidth}
        \includegraphics[width=\textwidth]{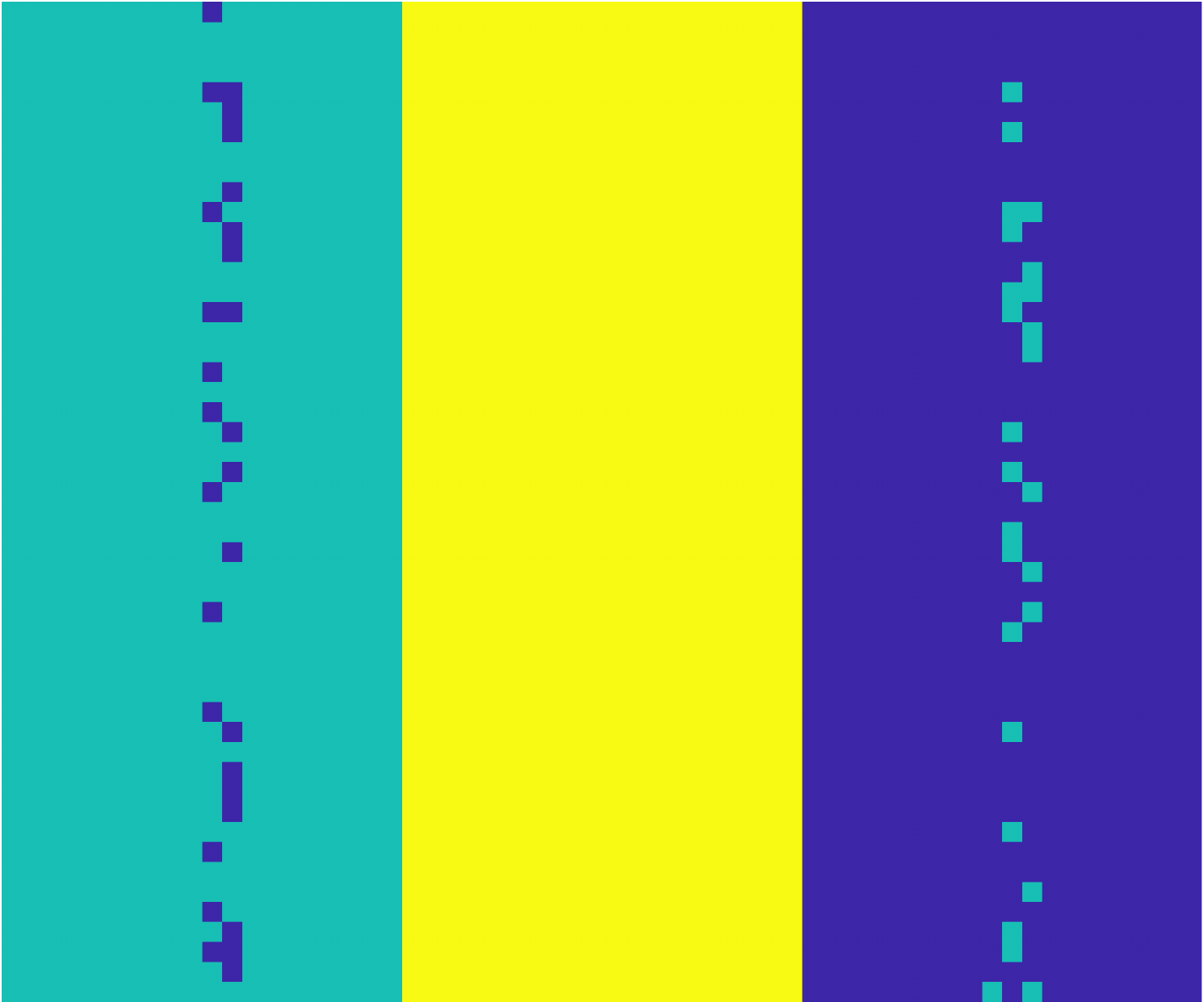}
        \subcaption{SC}
    \end{subfigure}
     \begin{subfigure}{0.09\textwidth}
        \includegraphics[width=\textwidth]{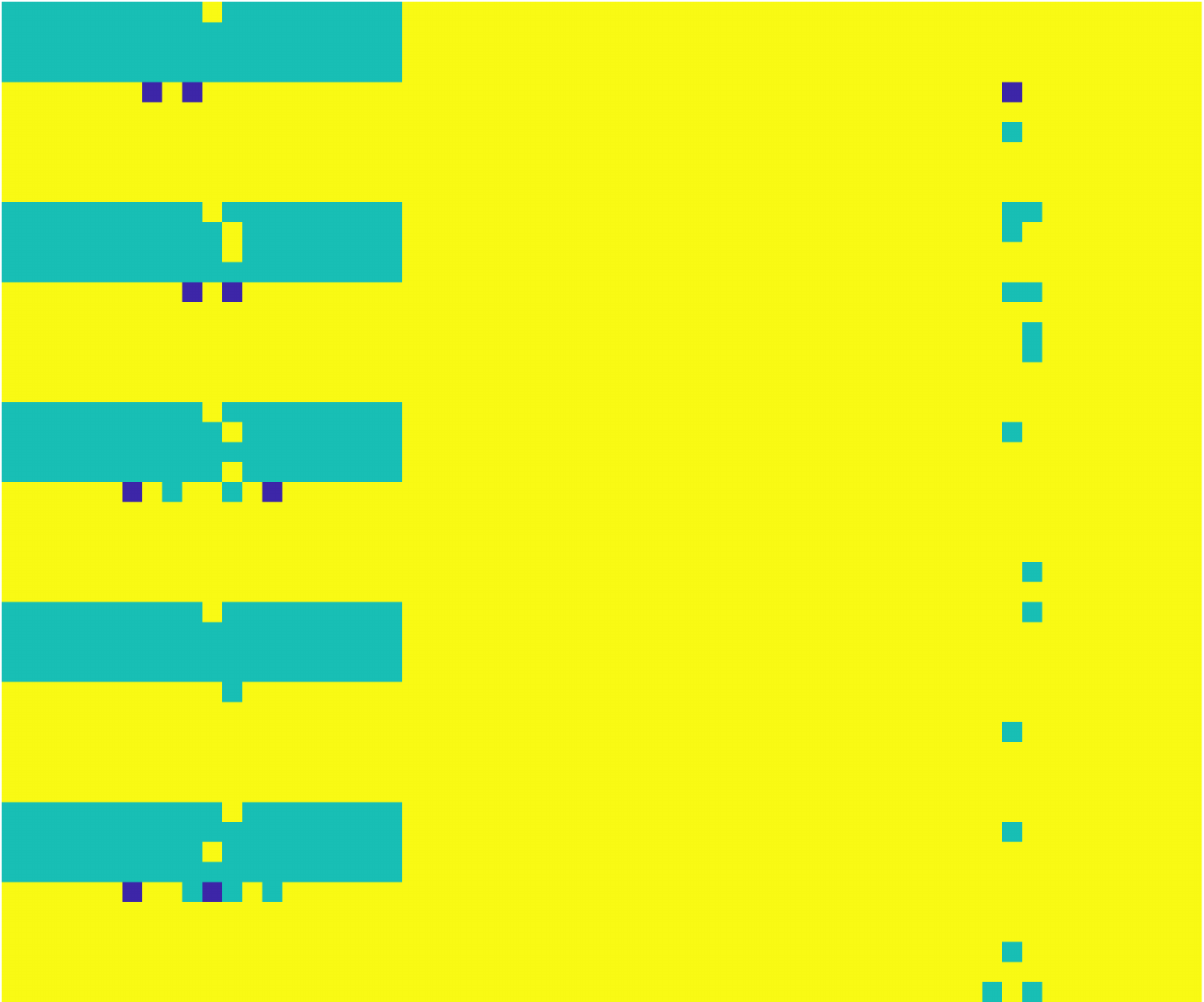}
        \subcaption{DL}
    \end{subfigure}
     \begin{subfigure}{0.09\textwidth}
        \includegraphics[width=\textwidth]{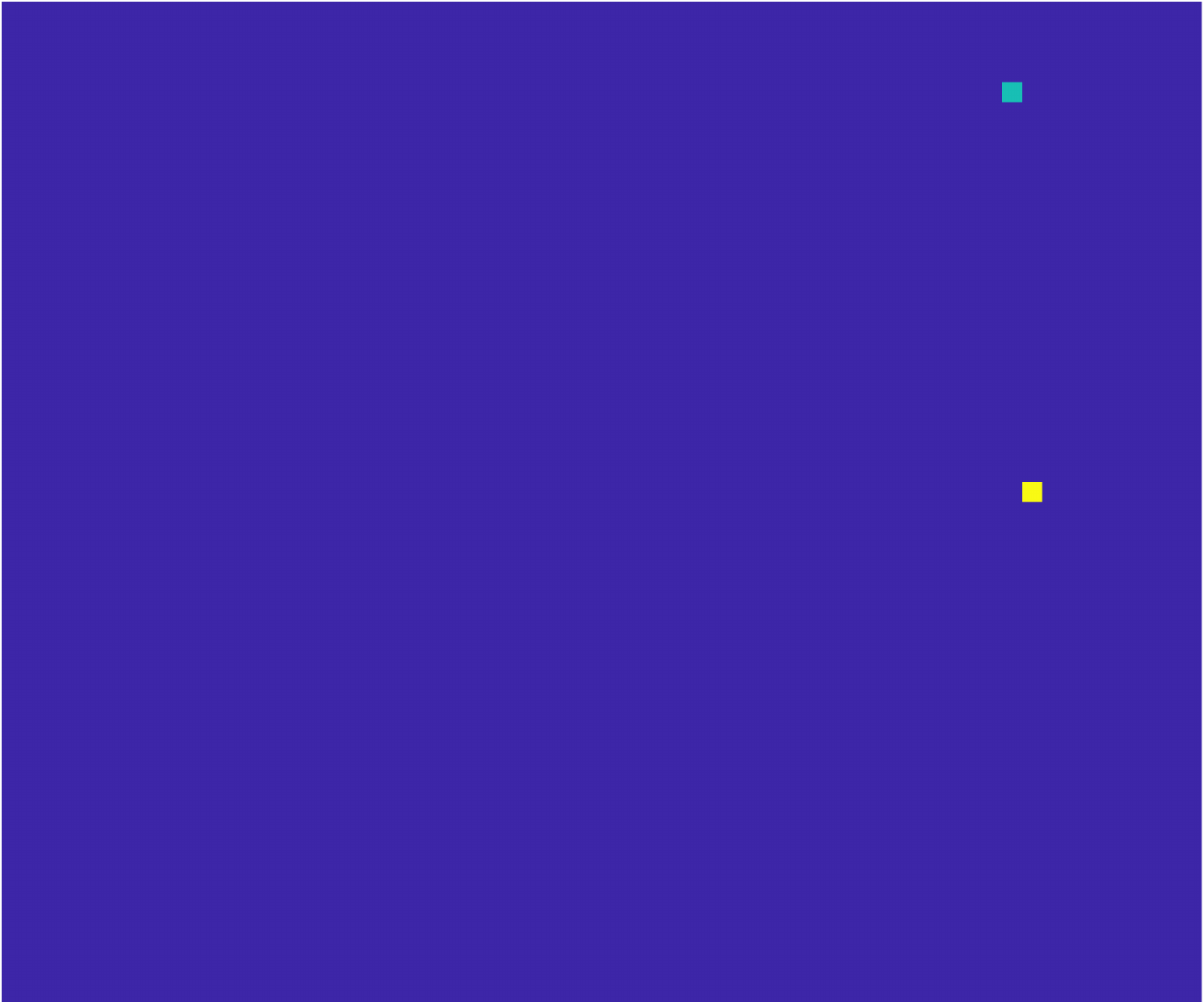}
        \subcaption{FSFDPC}
    \end{subfigure}
     \begin{subfigure}{0.09\textwidth}
        \includegraphics[width=\textwidth]{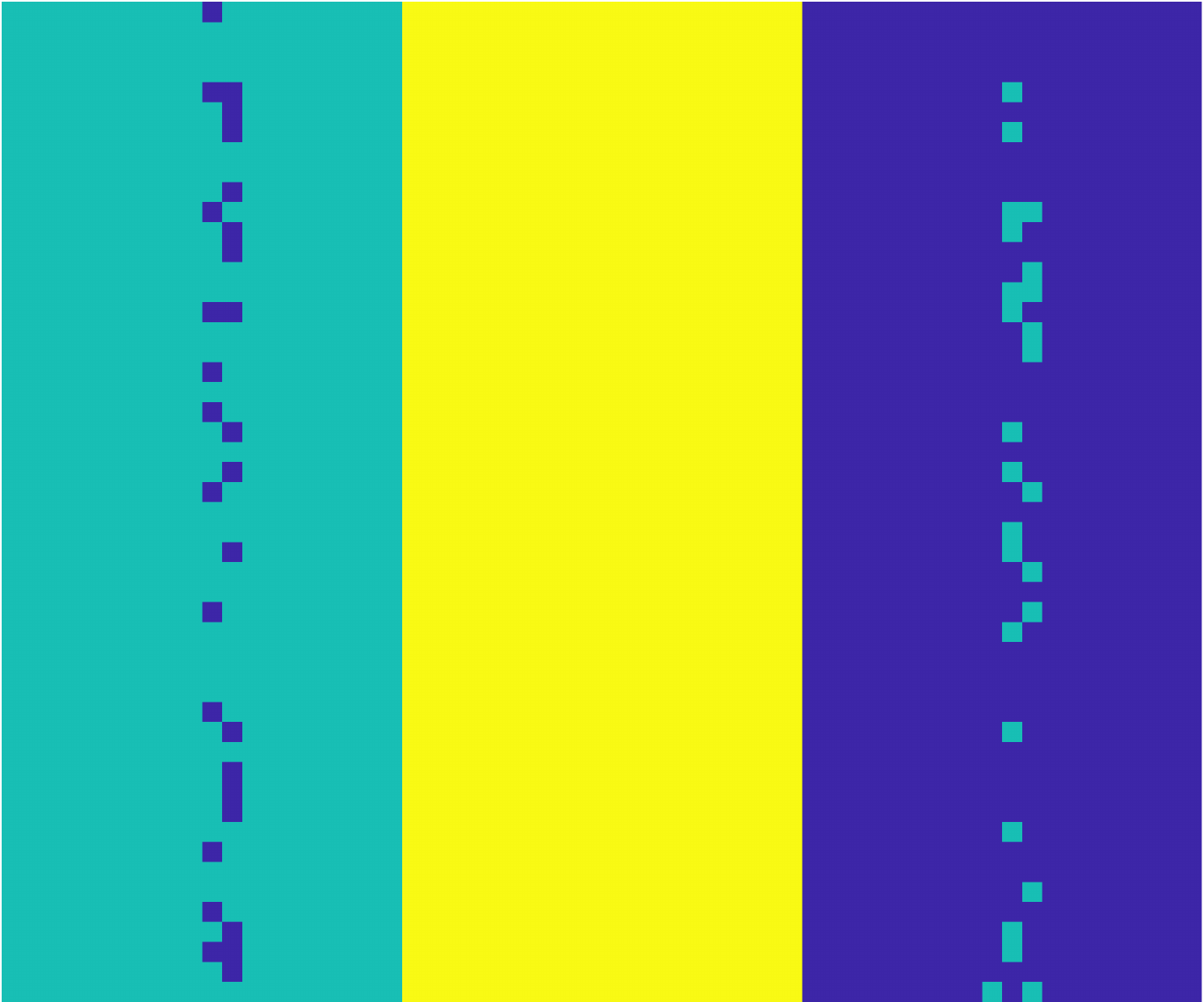}
        \subcaption{NMF}
    \end{subfigure}
     \begin{subfigure}{0.09\textwidth}
        \includegraphics[width=\textwidth]{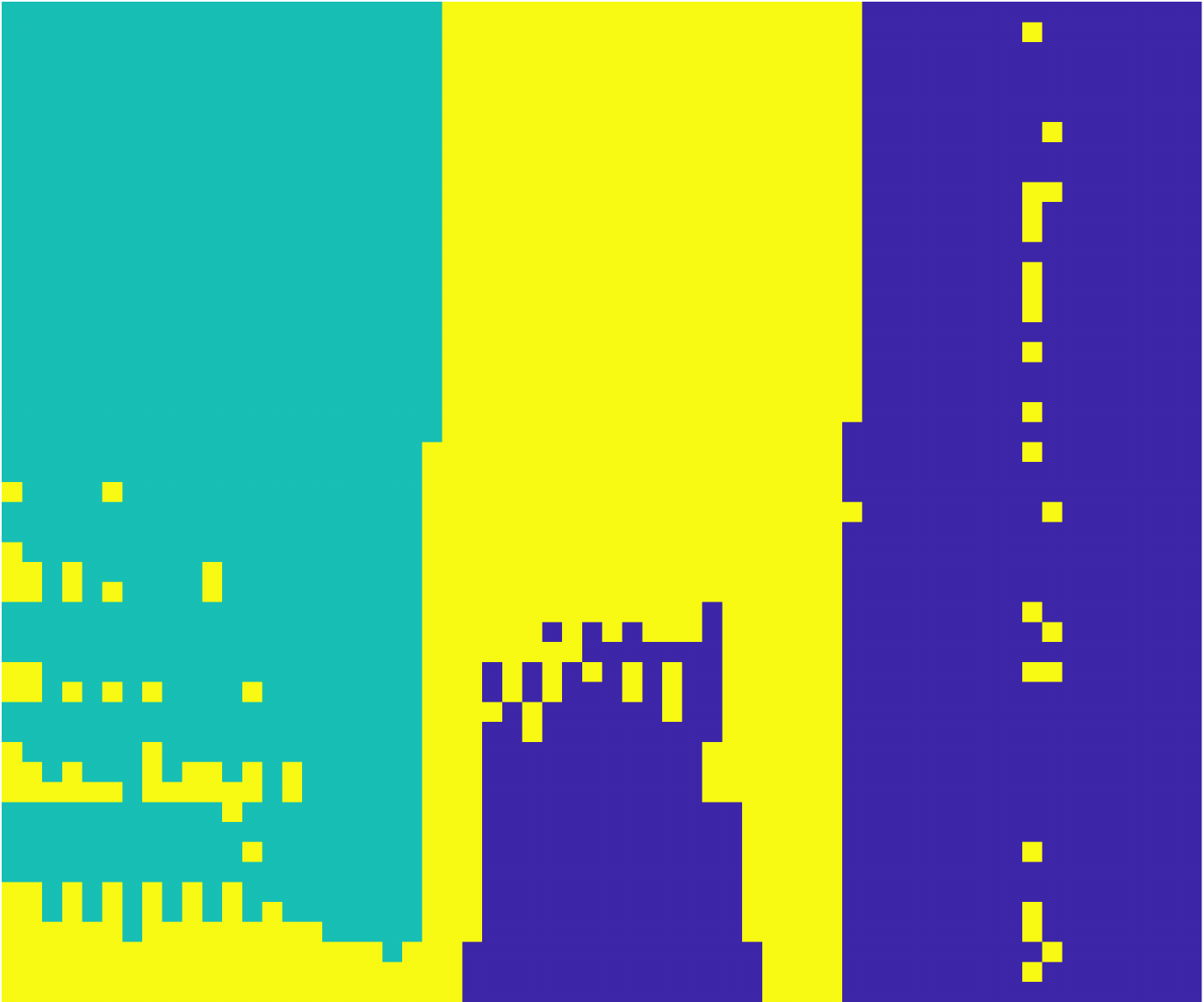}
        \subcaption{LCMR}
    \end{subfigure}
     \begin{subfigure}{0.09\textwidth}
        \includegraphics[width=\textwidth]{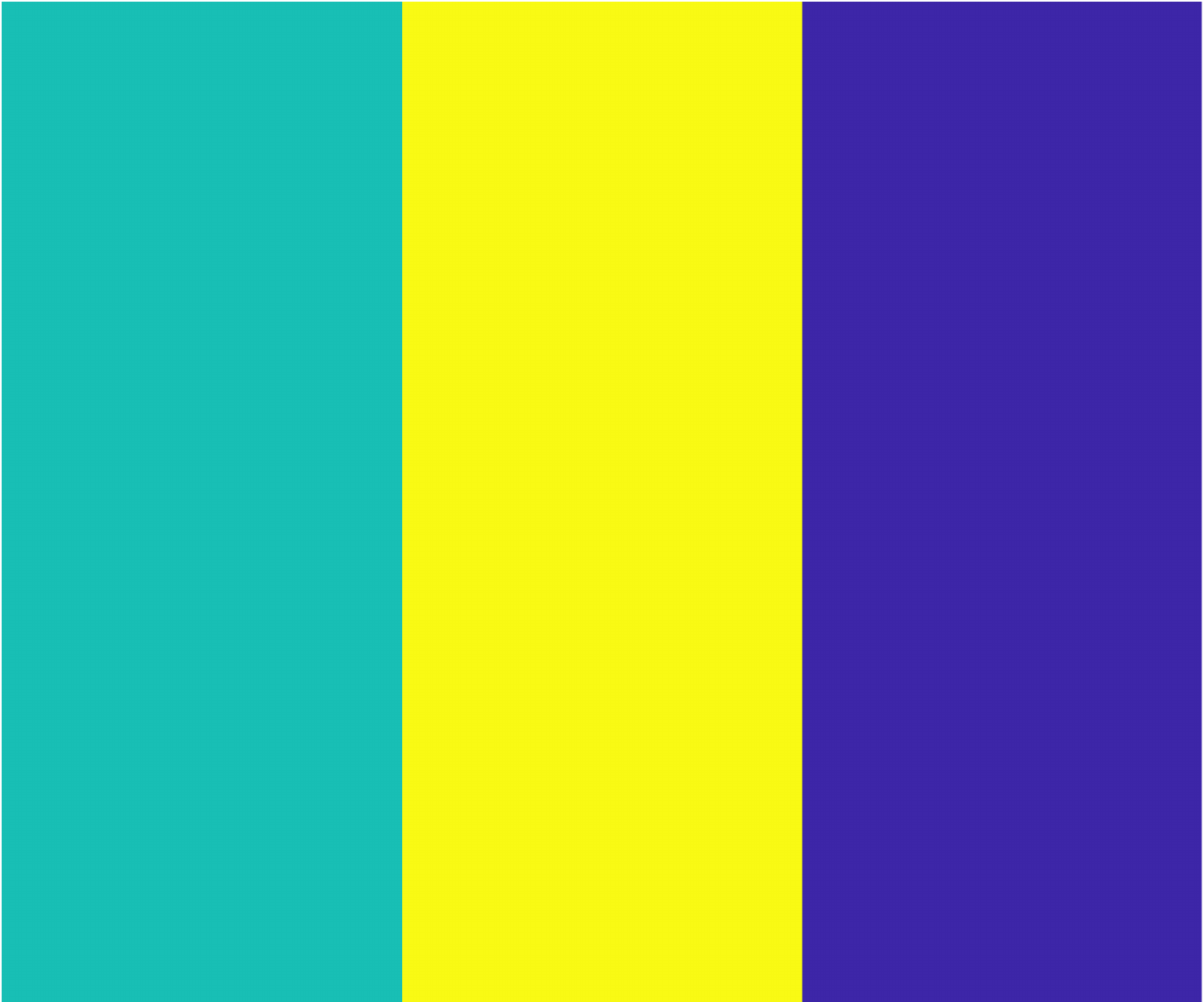}
        \subcaption{SRUSC}
    \end{subfigure}
    \caption{\label{fig:ThreeCubes_Results}  On the synthetic three cubes data set, only the proposed method is able to correctly label all data points.   In particular, the spatial regularization is necessary to gain robustness to the noise introduced.}
    \end{figure}

\begin{figure}[!htb]
    \centering
     \begin{subfigure}{0.09\textwidth}
        \includegraphics[width=\textwidth]{images/SalinasA/SalinasA_gt-crop.pdf}
        \subcaption{GT}
    \end{subfigure}
    \begin{subfigure}{0.09\textwidth}
        \includegraphics[width=\textwidth]{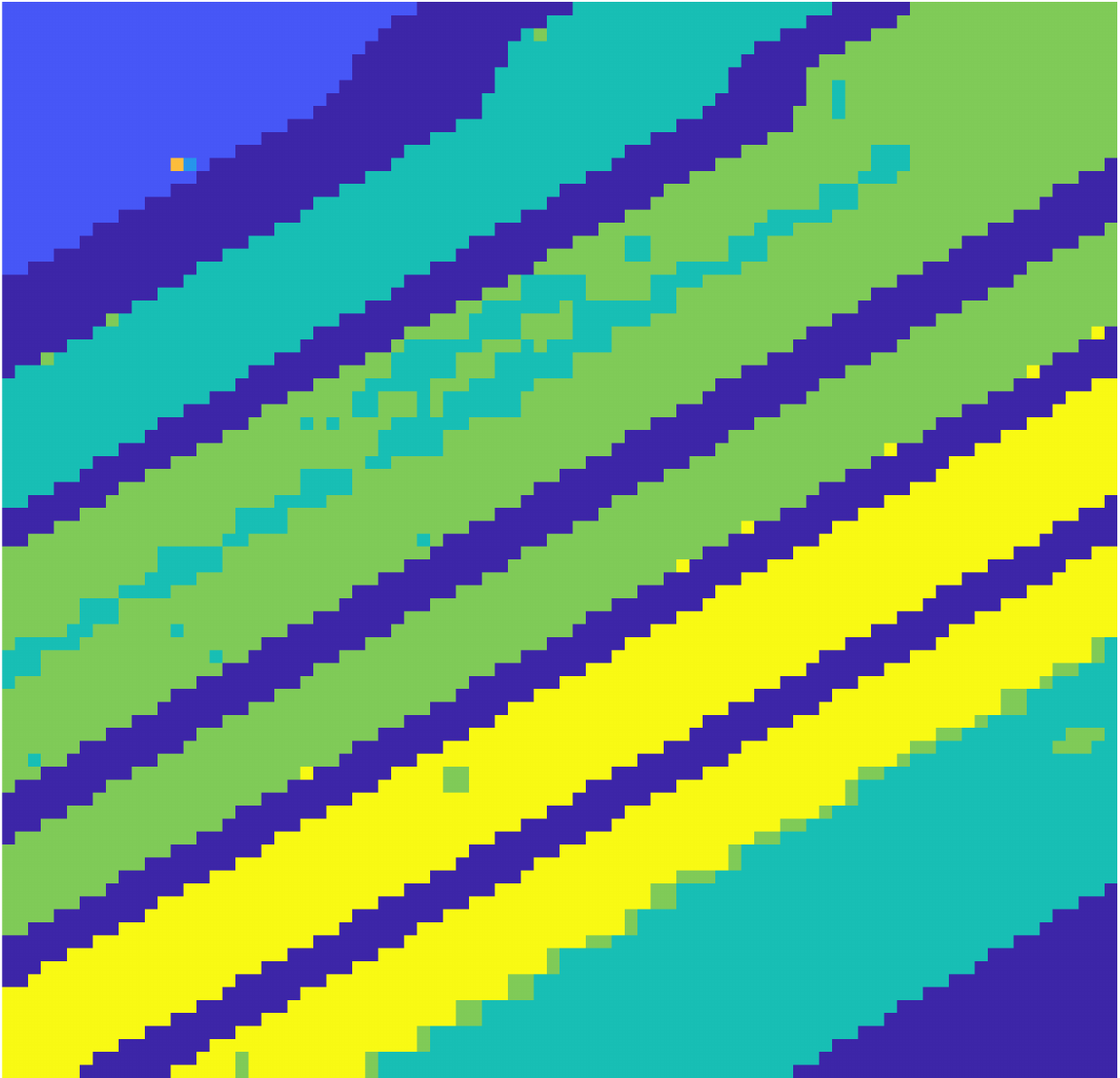}
        \subcaption{KM}
    \end{subfigure}
    \begin{subfigure}{0.09\textwidth}
        \includegraphics[width=\textwidth]{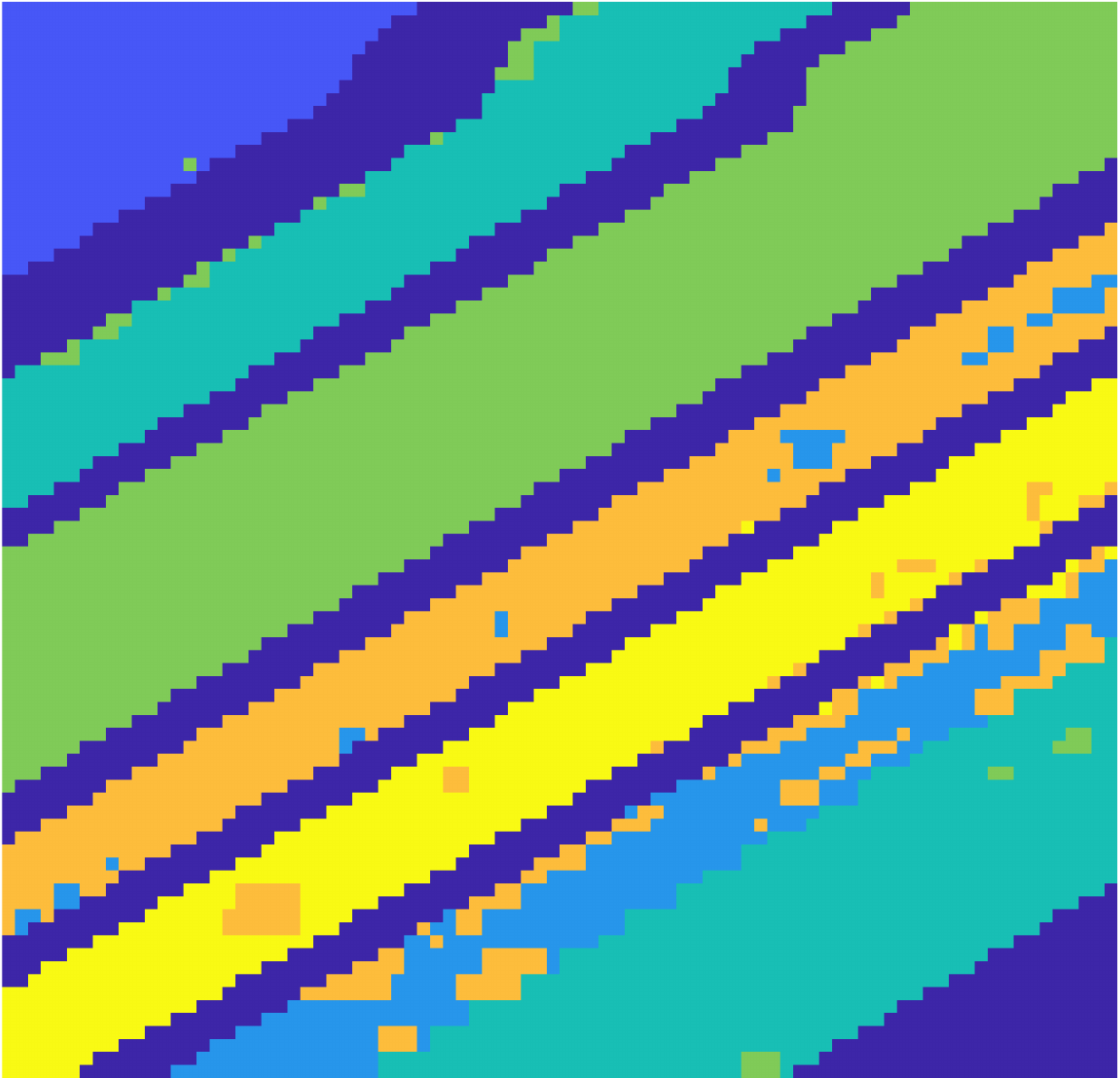}
        \subcaption{PCA}
    \end{subfigure}
    \begin{subfigure}{0.09\textwidth}
        \includegraphics[width=\textwidth]{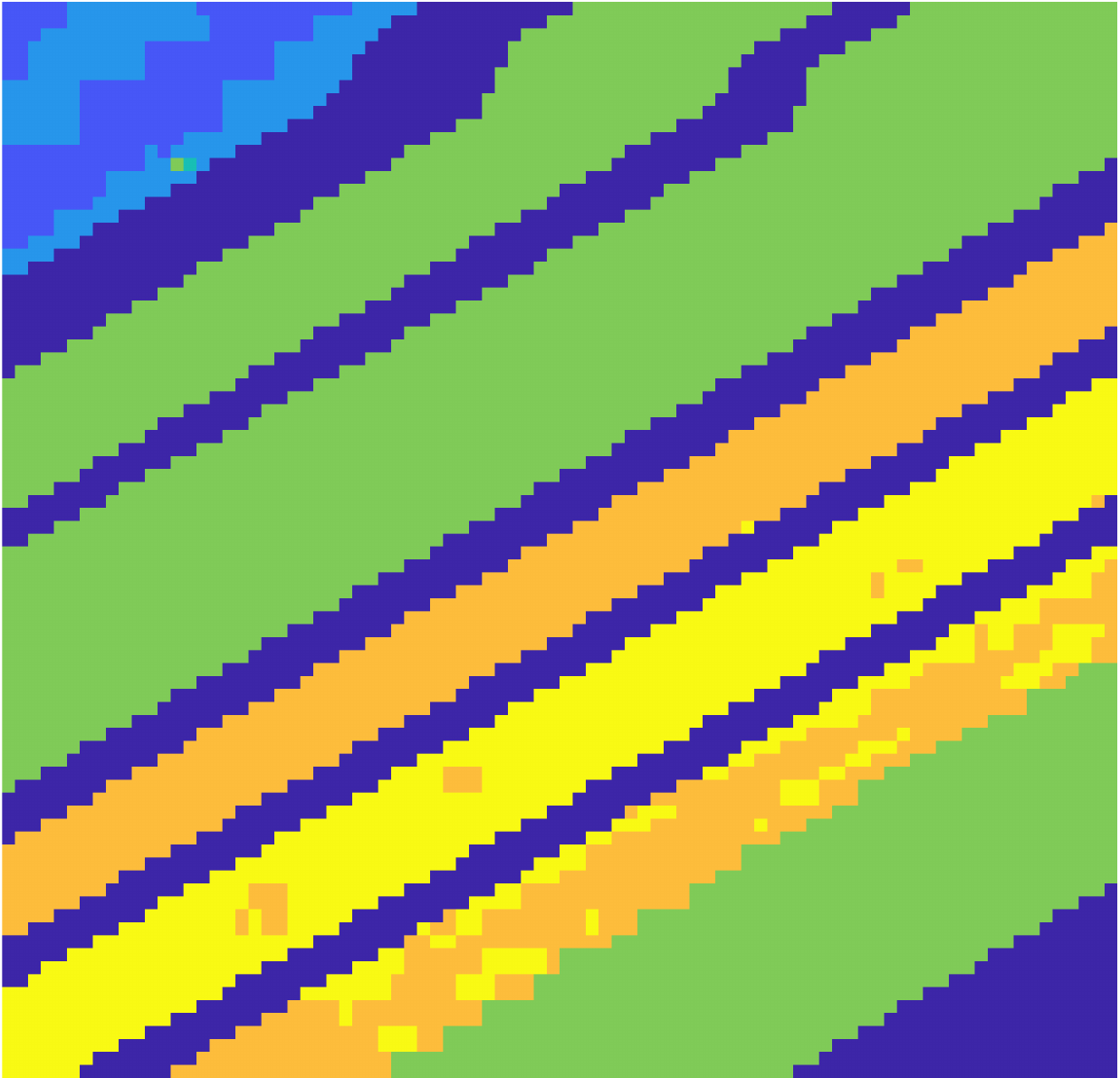}
        \subcaption{GMM}
    \end{subfigure}
     \begin{subfigure}{0.09\textwidth}
        \includegraphics[width=\textwidth]{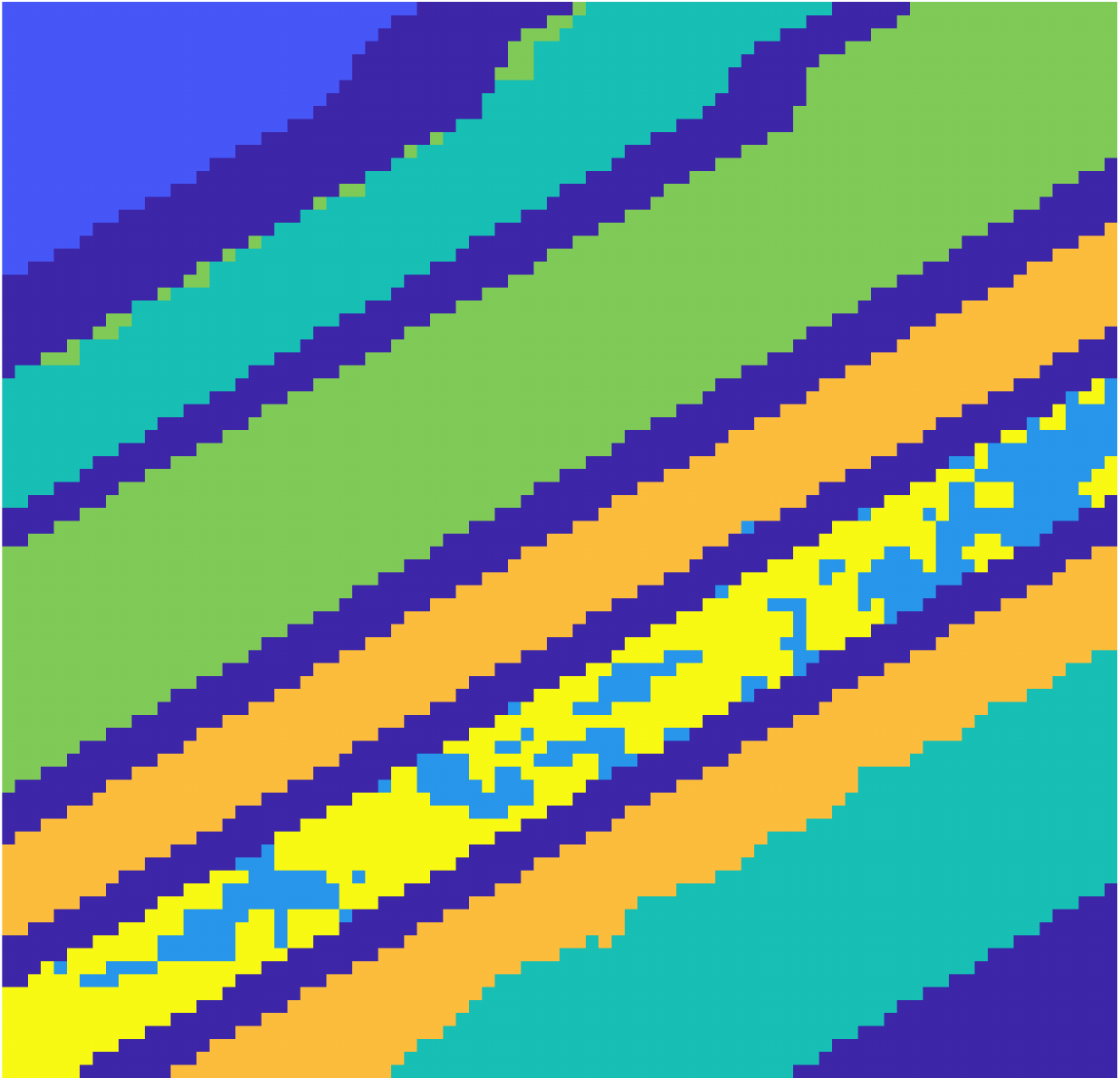}
        \subcaption{SC}
    \end{subfigure}
     \begin{subfigure}{0.09\textwidth}
        \includegraphics[width=\textwidth]{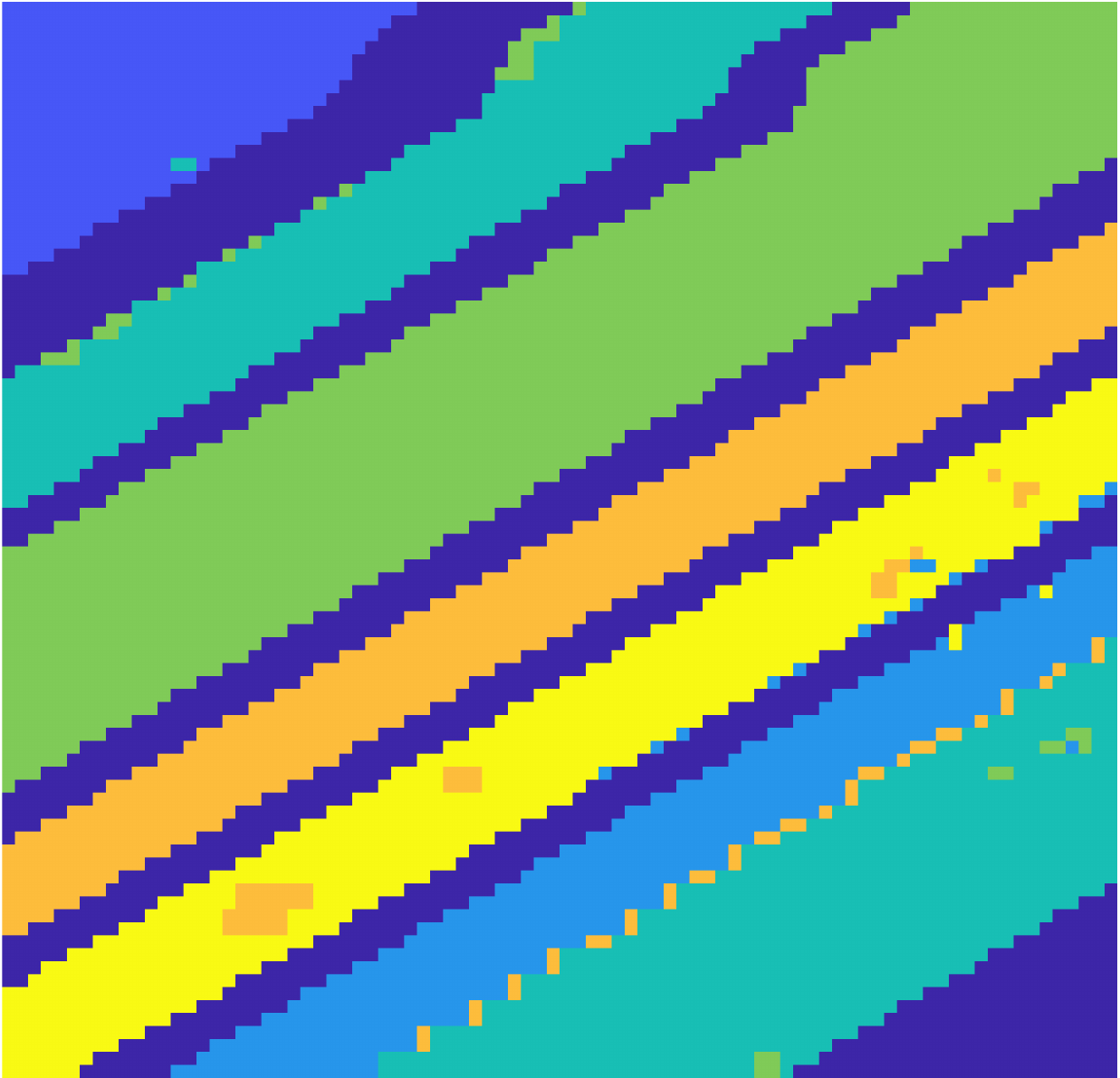}
        \subcaption{DL}
    \end{subfigure}
     \begin{subfigure}{0.09\textwidth}
        \includegraphics[width=\textwidth]{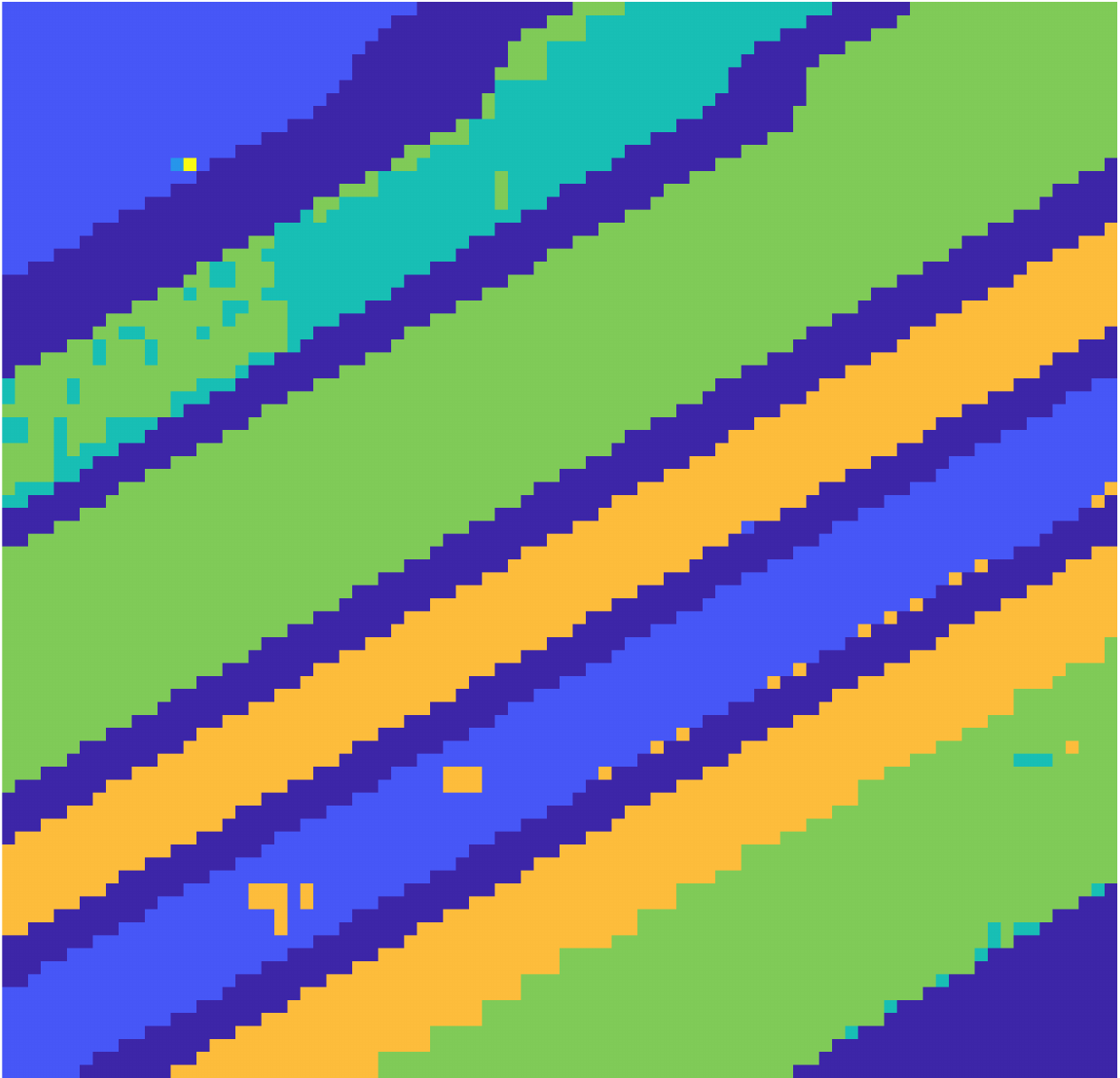}
        \subcaption{FSFDPC}
    \end{subfigure}
     \begin{subfigure}{0.09\textwidth}
        \includegraphics[width=\textwidth]{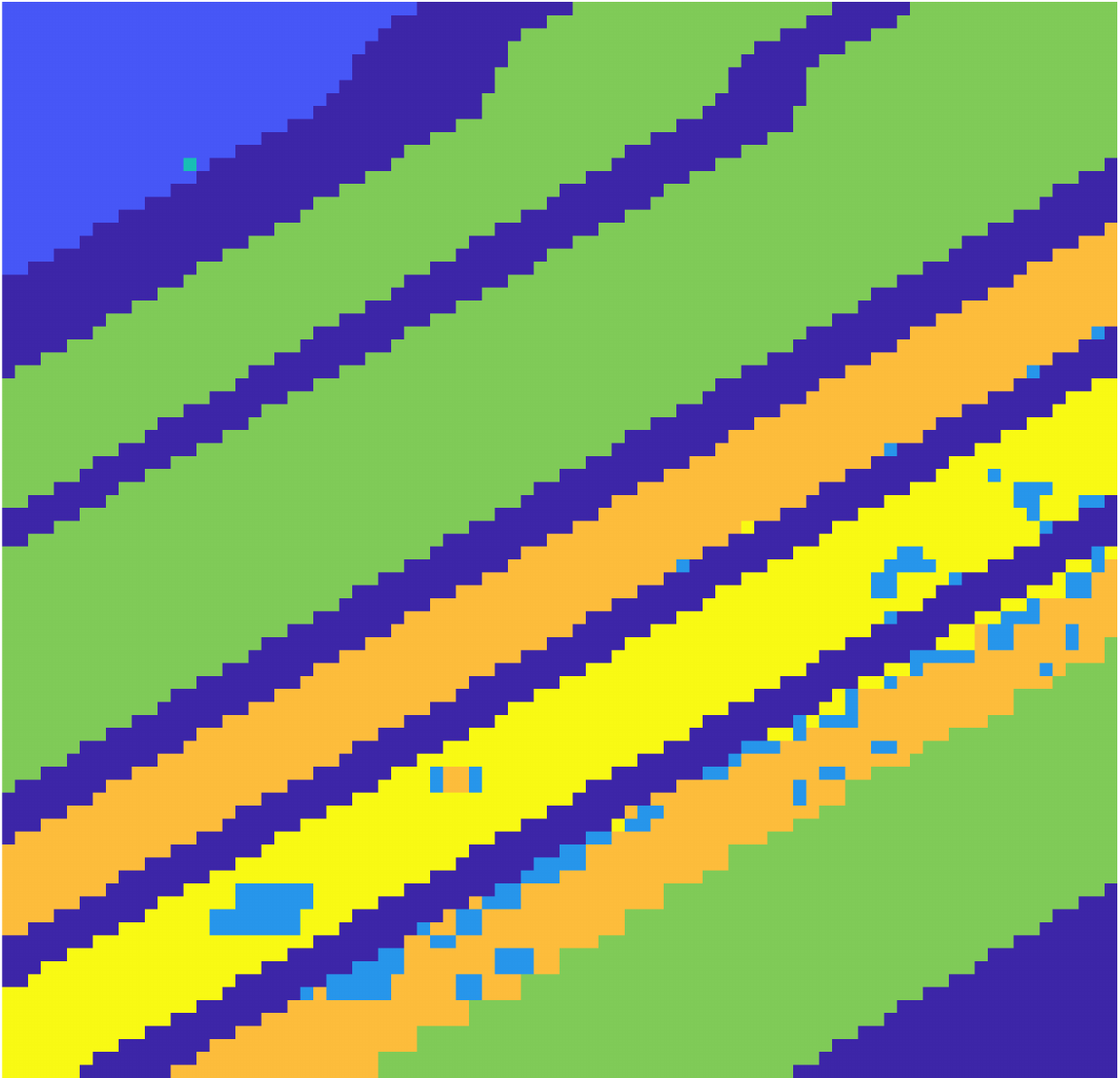}
        \subcaption{NMF}
    \end{subfigure}
     \begin{subfigure}{0.09\textwidth}
        \includegraphics[width=\textwidth]{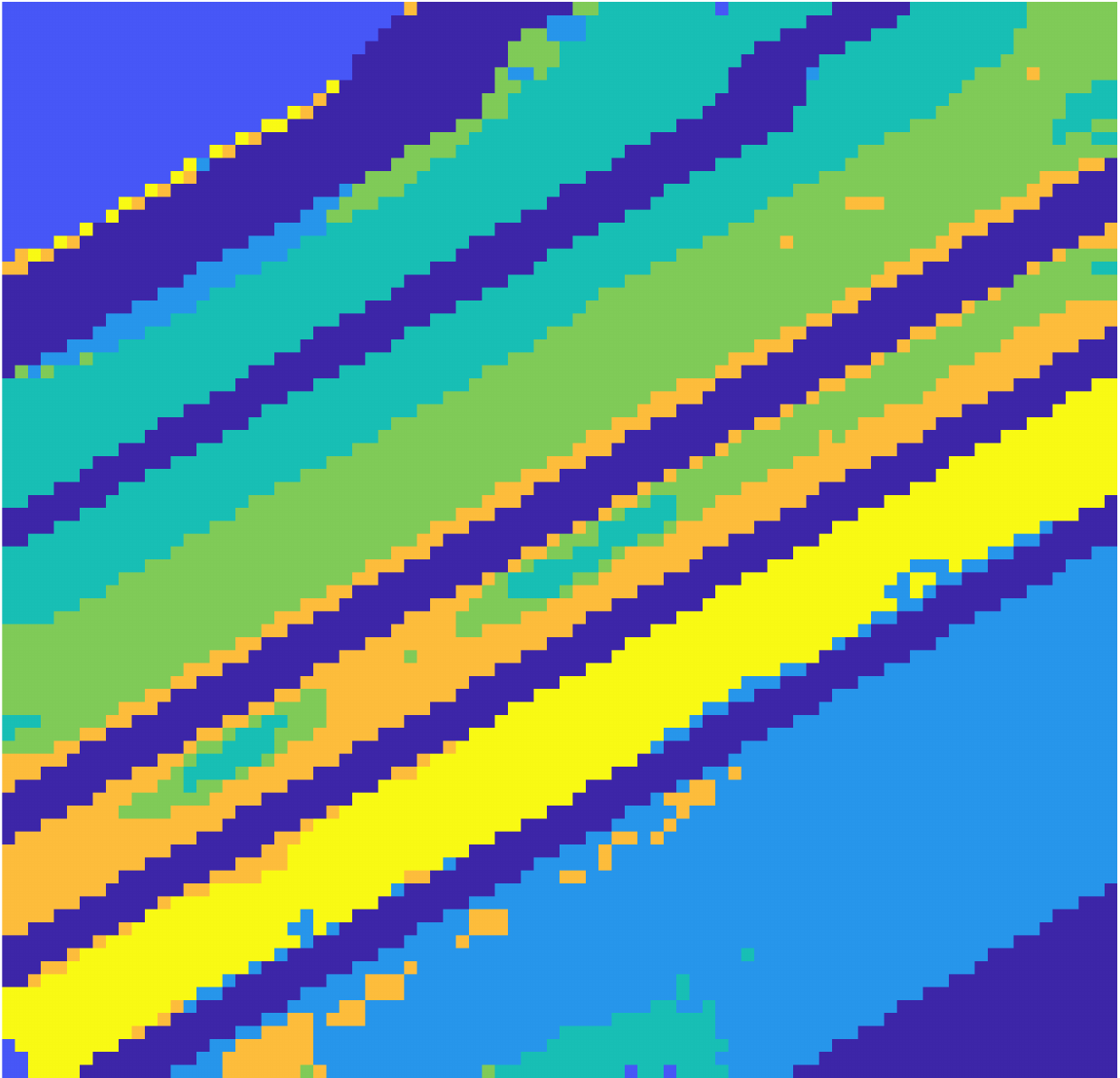}
        \subcaption{LCMR}
    \end{subfigure}
     \begin{subfigure}{0.09\textwidth}
        \includegraphics[width=\textwidth]{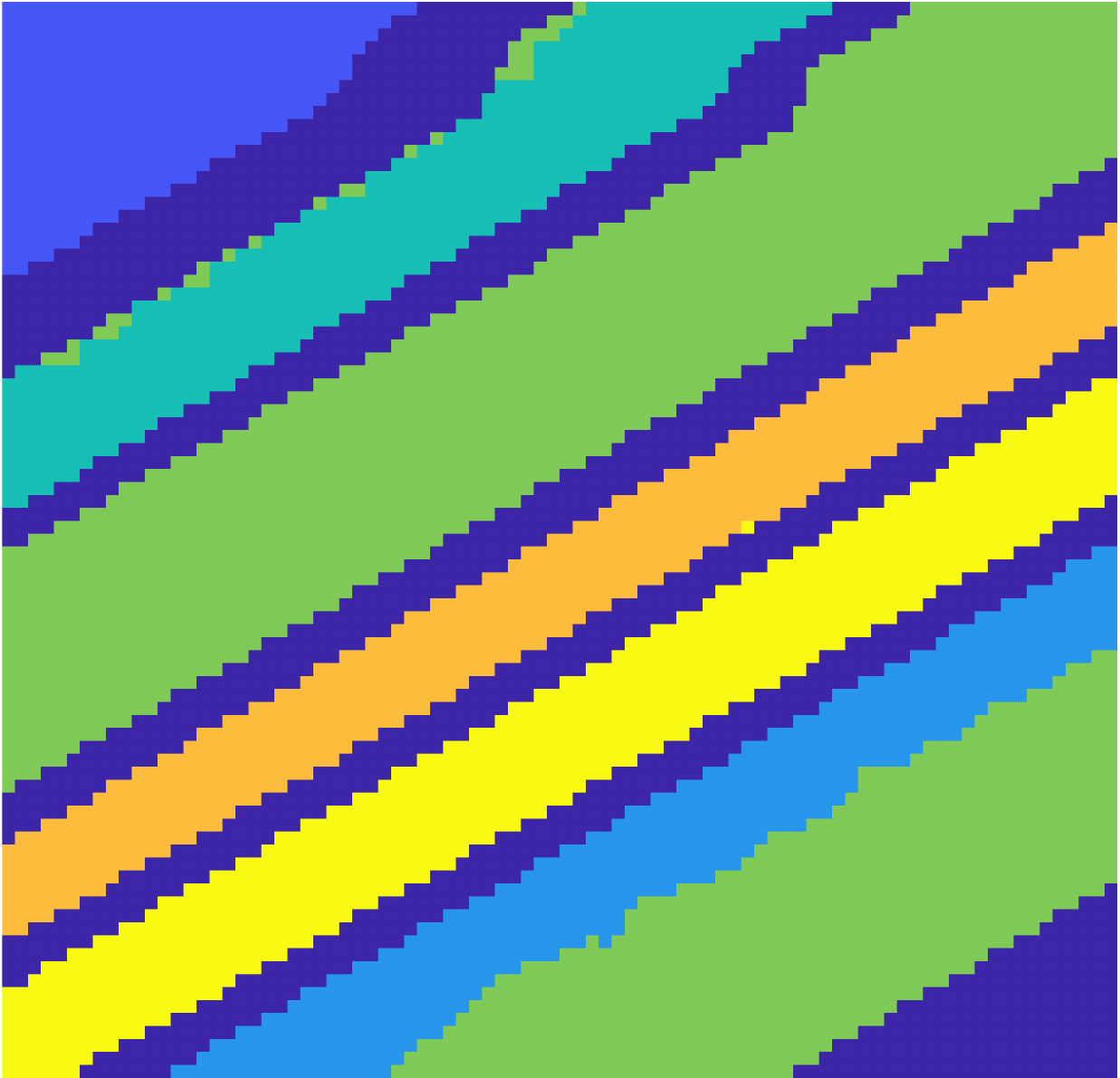}
        \subcaption{SRUSC}
    \end{subfigure}
    \caption{\label{fig:SalinasA_Results}  On the Salinas A data set, the proposed method performs strongly, with DL also performing well.}
    \end{figure}

\begin{figure}[!htb]
    \centering
     \begin{subfigure}{0.09\textwidth}
        \includegraphics[width=\textwidth]{images/PaviaU/PaviaU_gt-crop.pdf}
        \subcaption{GT}
    \end{subfigure}
    \begin{subfigure}{0.09\textwidth}
        \includegraphics[width=\textwidth]{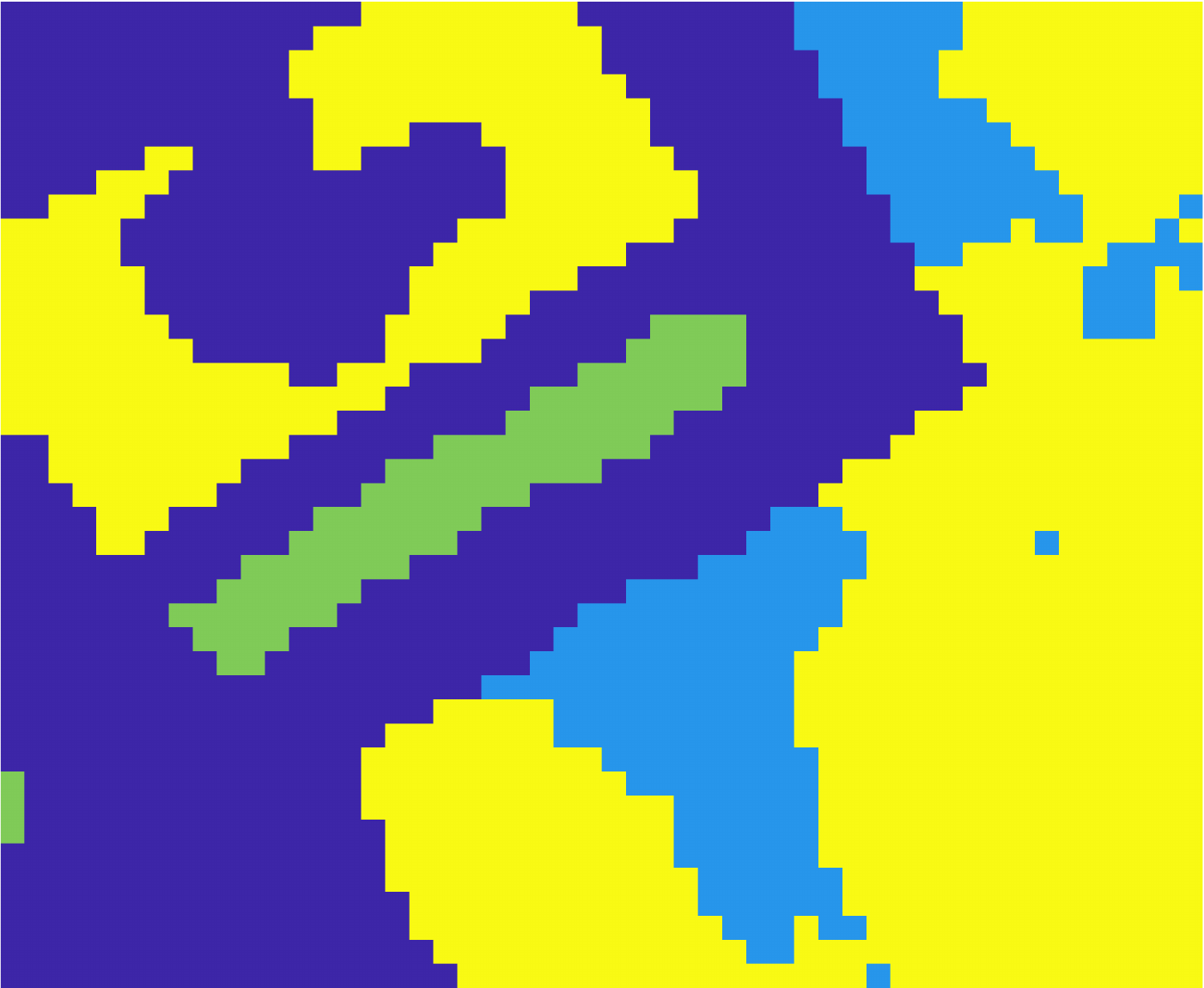}
        \subcaption{KM}
    \end{subfigure}
    \begin{subfigure}{0.09\textwidth}
        \includegraphics[width=\textwidth]{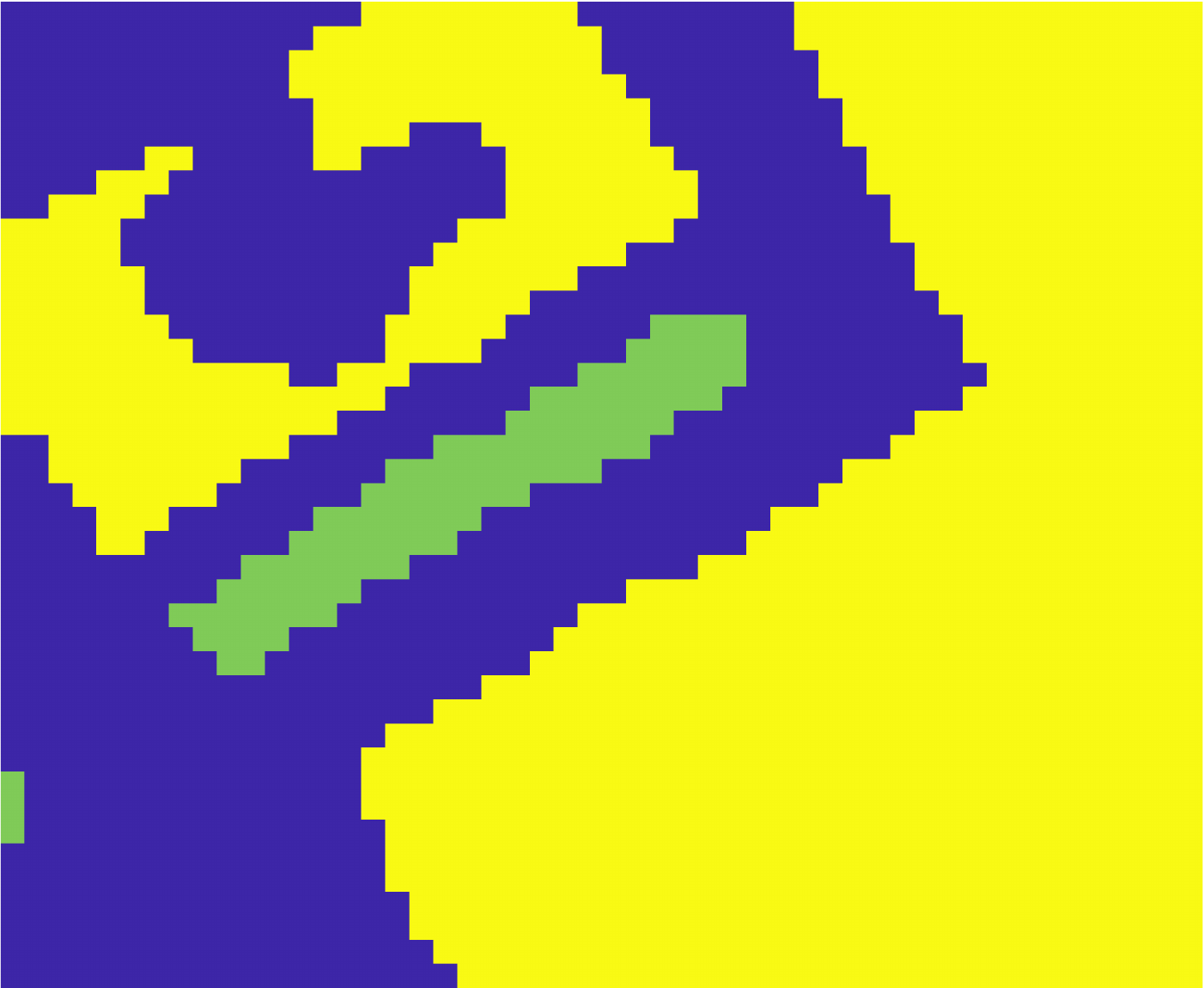}
        \subcaption{PCA}
    \end{subfigure}
    \begin{subfigure}{0.09\textwidth}
        \includegraphics[width=\textwidth]{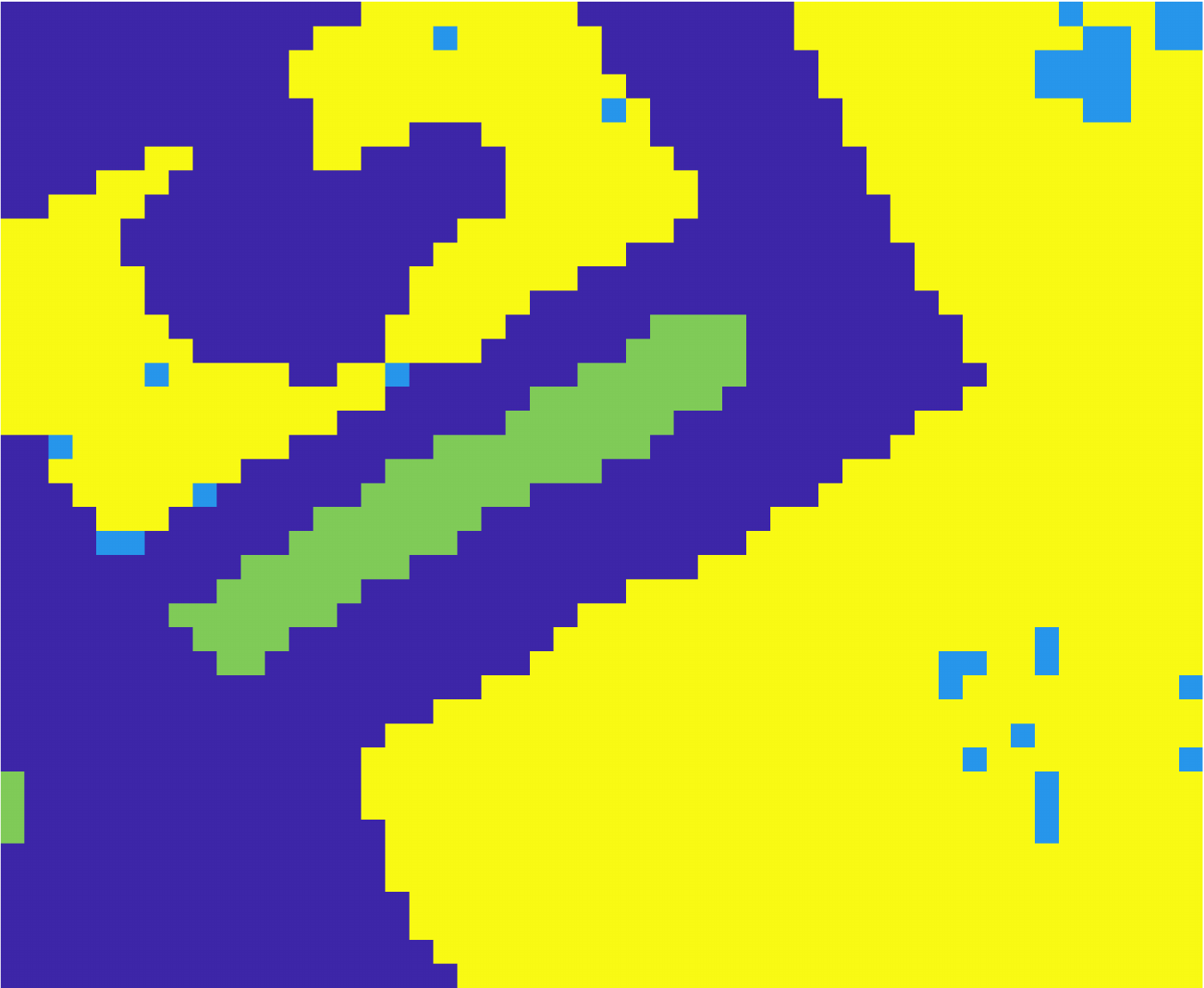}
        \subcaption{GMM}
    \end{subfigure}
     \begin{subfigure}{0.09\textwidth}
        \includegraphics[width=\textwidth]{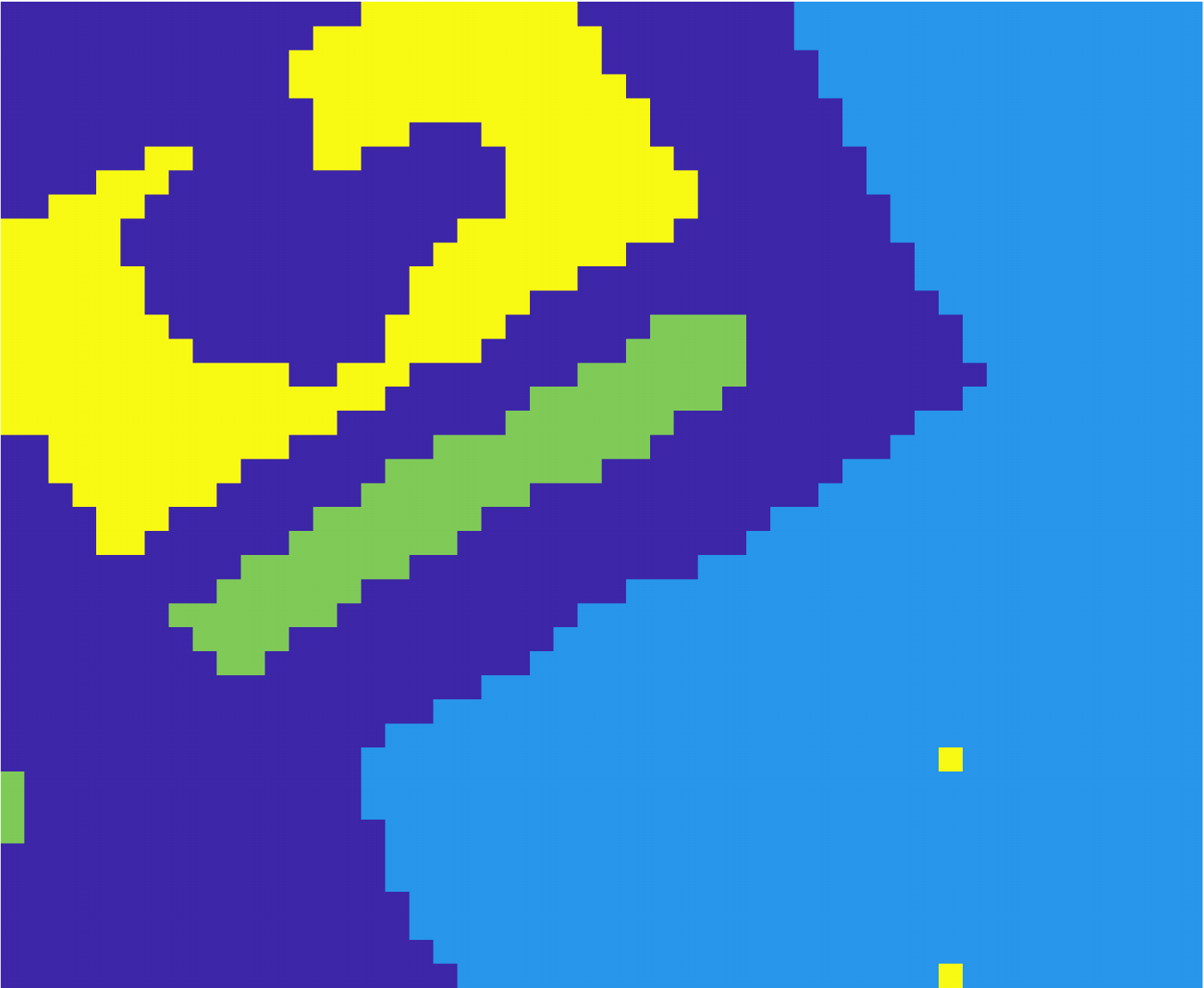}
        \subcaption{SC}
    \end{subfigure}
     \begin{subfigure}{0.09\textwidth}
        \includegraphics[width=\textwidth]{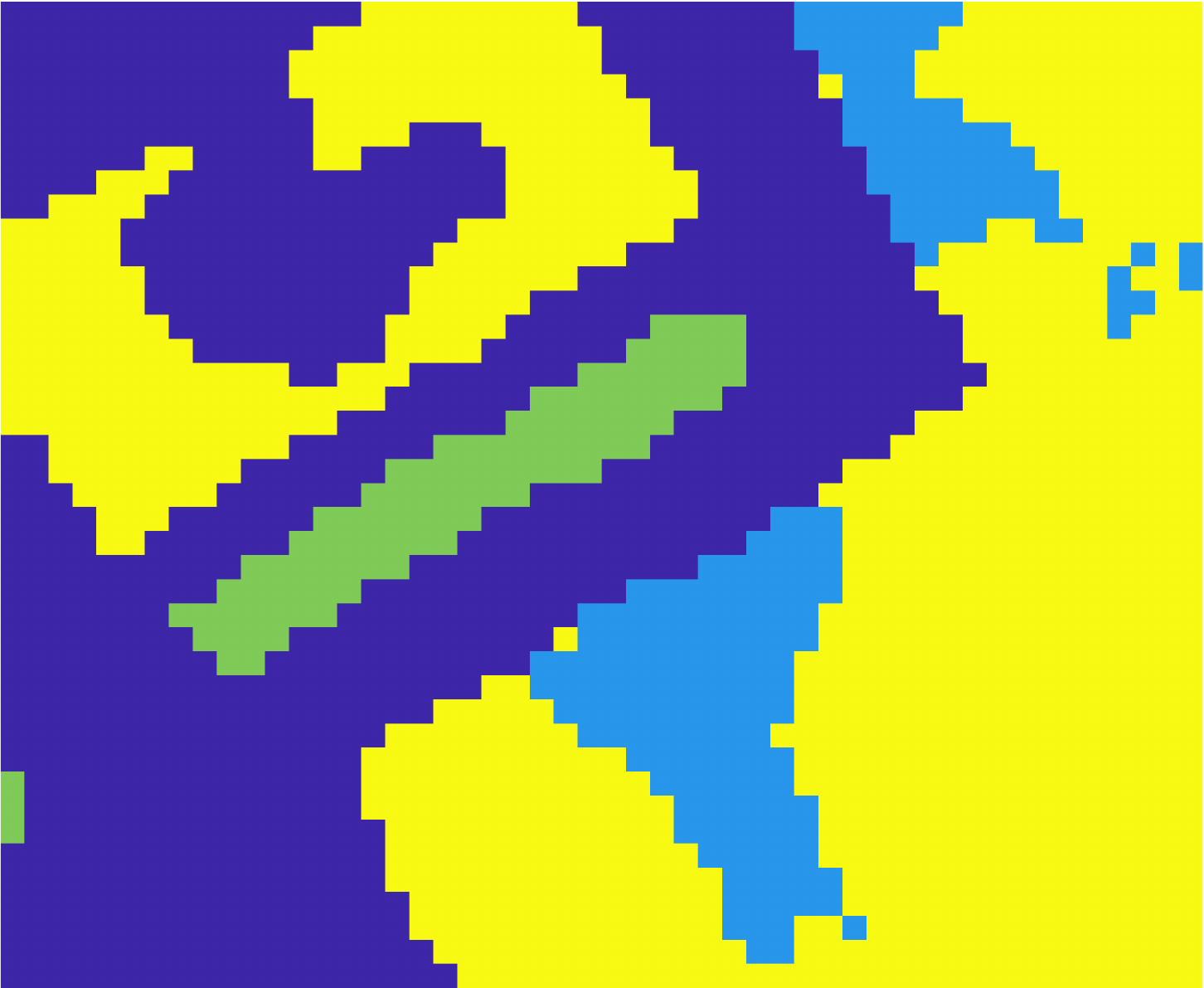}
        \subcaption{DL}
    \end{subfigure}
     \begin{subfigure}{0.09\textwidth}
        \includegraphics[width=\textwidth]{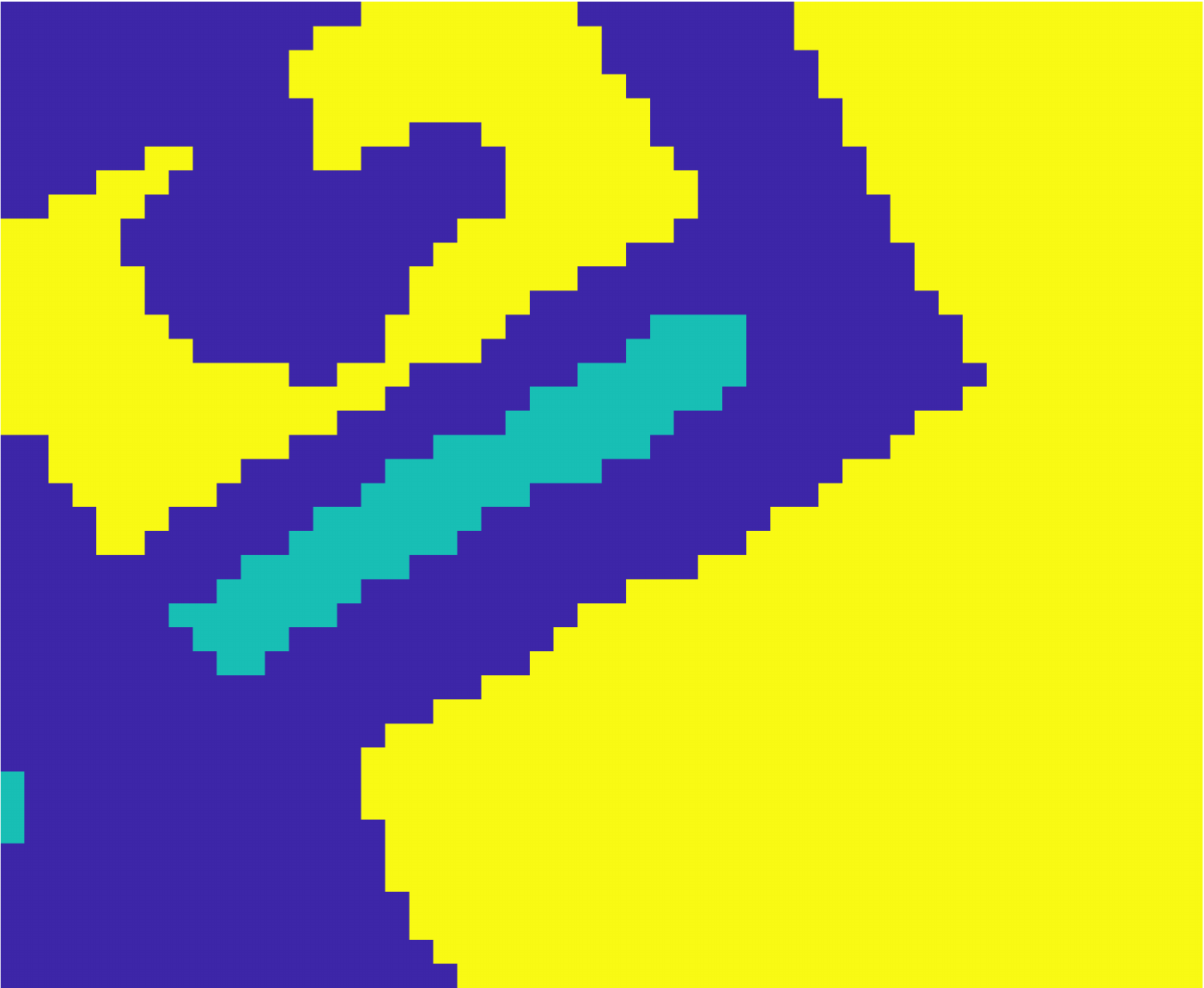}
        \subcaption{FSFDPC}
    \end{subfigure}
     \begin{subfigure}{0.09\textwidth}
        \includegraphics[width=\textwidth]{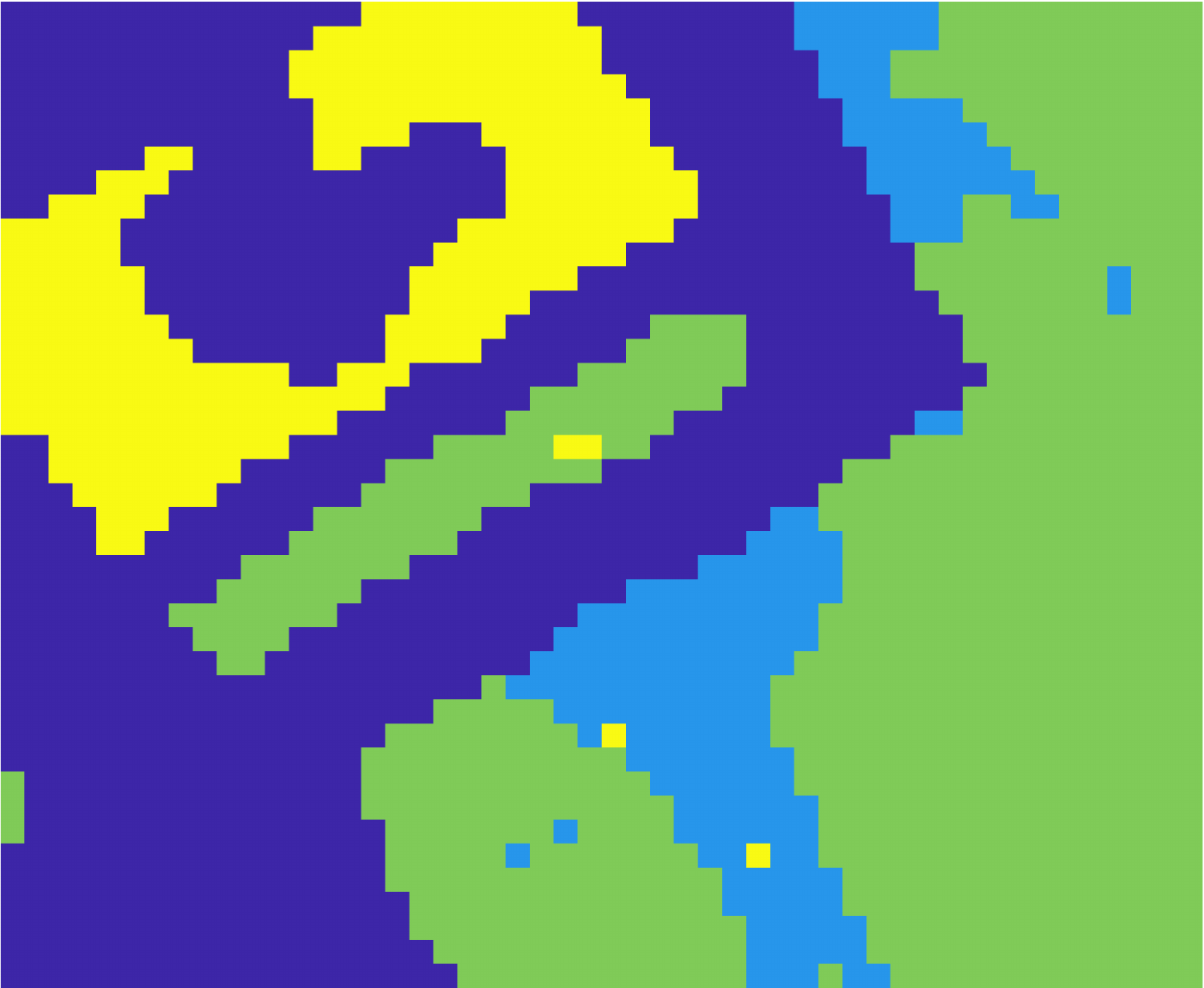}
        \subcaption{NMF}
    \end{subfigure}
     \begin{subfigure}{0.09\textwidth}
        \includegraphics[width=\textwidth]{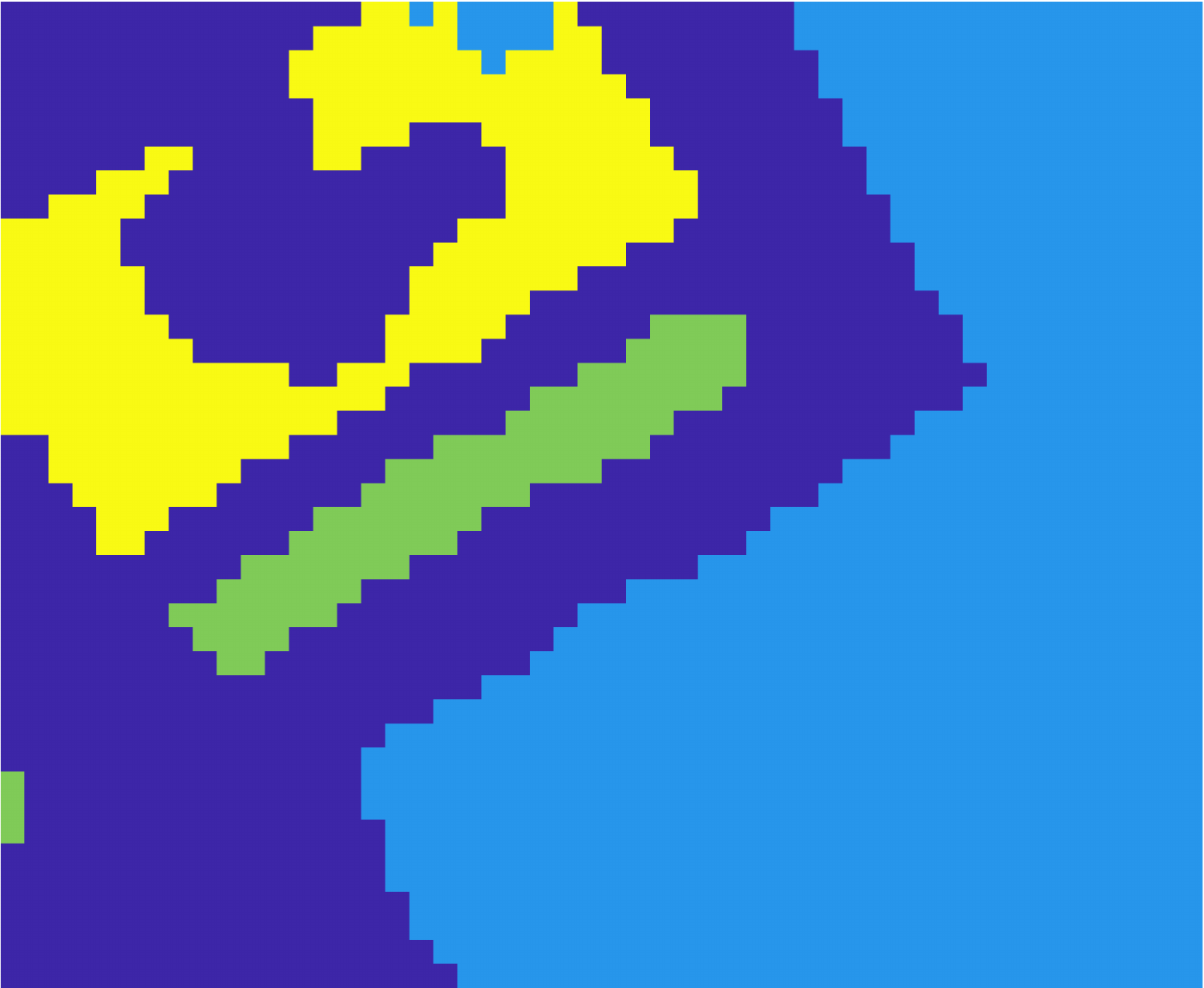}
        \subcaption{LCMR}
    \end{subfigure}
     \begin{subfigure}{0.09\textwidth}
        \includegraphics[width=\textwidth]{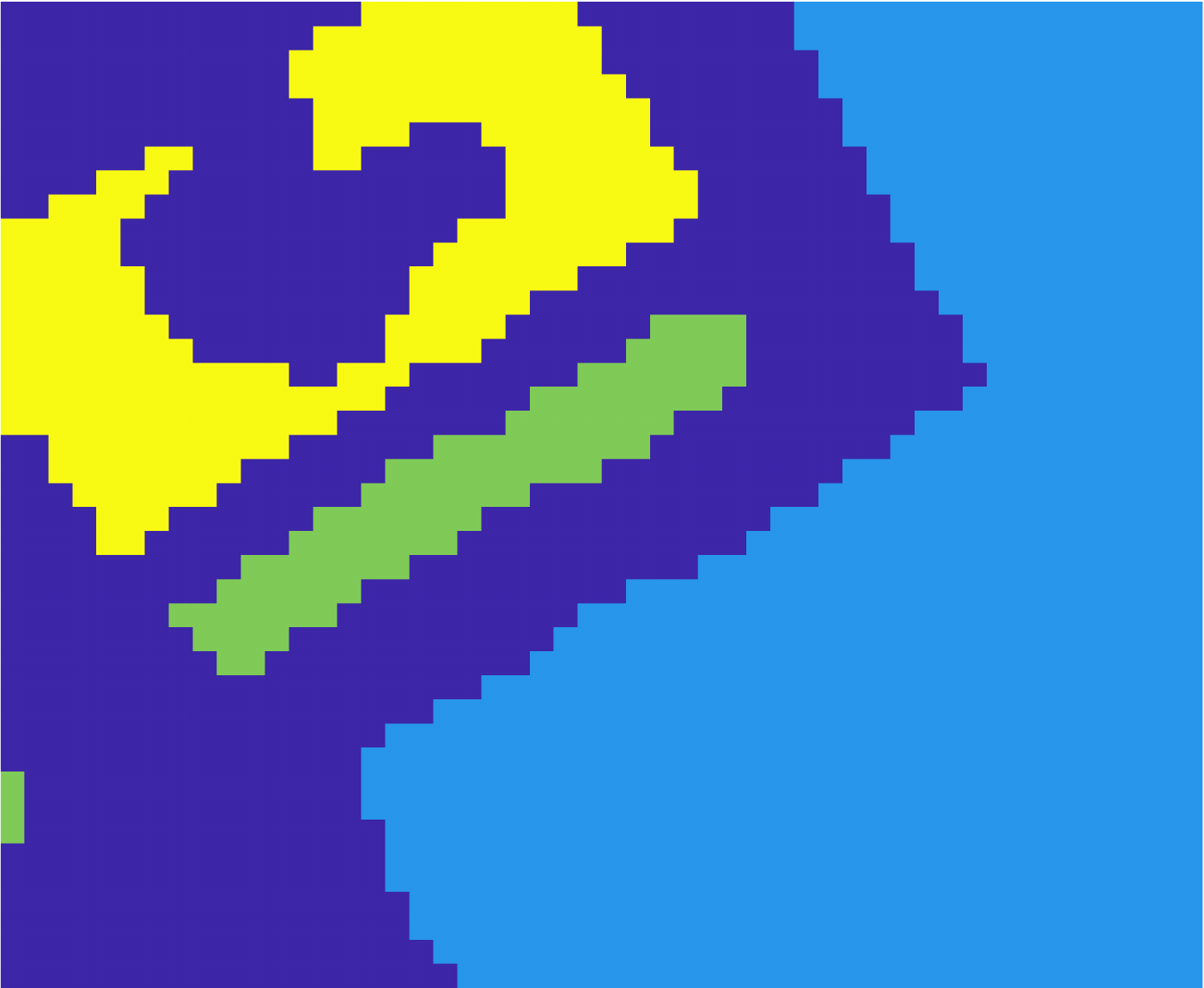}
        \subcaption{SRUSC}
    \end{subfigure}
    \caption{\label{fig:PaviaU_Results}On the Pavia U data set, only the proposed method achieves perfect accuracy, though Euclidean spectral clustering and LCMR perform well.}
    \end{figure}

To perform quantitative comparisons, we align each clustered data set with the ground truth by solving a linear assignment problem with the Hungarian algorithm, then computing the overall accuracy (OA), average accuracy (AA), and Cohen's $\kappa$ statistic for each clustering.  Numerical results are in Table \ref{tab:Results}, with visual results in Fig. \ref{fig:FourCircles_Results},\ref{fig:ThreeCubes_Results},\ref{fig:SalinasA_Results},\ref{fig:PaviaU_Results}.

\begin{table*}[htb!]
\centering
\begin{adjustbox}{max width=\textwidth}
\begin{tabular}{ | c | c | c | c | c | c | c | c | c | c | c | c | c |  }
 \hline
 Data set & FS AA & FS OA & FS $\kappa$ & TC AA & TC OA & TC $\kappa$ & SalinasA AA & SalinasA OA & SalinasA $\kappa$& PaviaU AA & PaviaU OA & PaviaU $\kappa$ \\
 \hline
KM  &   \textbf{1.0000} & \textbf{1.0000} & \textbf{1.0000} & \underline{0.9817} & \underline{0.9817}  & \underline{0.9725} & 0.6577  &   0.6260  &   0.5256  &   0.5787  &   0.6041  &   0.0857  \\
PCA     &   \textbf{1.0000} & \textbf{1.0000} & \textbf{1.0000} & \underline{0.9817} & \underline{0.9817}  & \underline{0.9725} & 0.8525   &   0.7992  &   0.7552  &   0.6667  &   0.7919  &   0.3868   \\
GMM     & 0.9347 & 0.9180 & 0.7990 & 0.6613 & 0.6613  & 0.4920 &   0.5819   &   0.5942  &   0.4805  &   0.7964  &   0.5675  &   0.3648  \\
SC      & \textbf{1.0000} & \textbf{1.0000} & \textbf{1.0000} & \underline{0.9817} & \underline{0.9817}  & \underline{0.9725} &   0.7235   &   0.7573  &   0.6941  &   \underline{0.9989}  &   \underline{0.9979}  &   \underline{0.9959}  \\
DL      & 0.7180 & 0.8590 & 0.5369 & 0.4657 & 0.4657  & 0.1985 &   \underline{0.8790}  &   \underline{0.8313}  &   \underline{0.7939}  &   0.3800  &   0.5927  &   0.0735  \\
FSFDPC  & \textbf{1.0000} & \textbf{1.0000} & \textbf{1.0000} & 0.3337 & 0.3337  & 0.0000 &   0.6055  &   0.6322  &   0.5377  &   0.3418  &   0.6528  &   0.0308   \\
NMF     & 0.9353 & 0.9030 & 0.7710 & \underline{0.9817} & \underline{0.9817}  & \underline{0.9725} &   0.6654  &   0.6402  &   0.5408  &   0.5935  &   0.7634  &   0.5196   \\
LCMR    & 0.5313 & 0.6470 & 0.0624 & 0.8123 & 0.8123  & 0.7185 &   0.7889  &   0.7620  &   0.7071  &   0.9870  &   0.9919  &   0.9815   \\
SRUSC     &  \textbf{1.0000} &  \textbf{1.0000} &  \textbf{1.0000} &  \textbf{1.0000} &  \textbf{1.0000}  &  \textbf{1.0000} &   \textbf{0.8866}  & \textbf{0.8545}  &  \textbf{0.8147}  & \textbf{1.0000} &  \textbf{1.0000} & \textbf{1.0000}\\
 \hline
\end{tabular}
\end{adjustbox}
\caption{\label{tab:Results}Results for clustering experiments.  We see that across all methods and data sets, the proposed SRUSC method gives the best clustering performance.  We note that several methods perform well on the synthetic four spheres data set: $K$-means, PCA followed by $K$-means, spectral clustering, and FSFDPC all give perfect performances.  However, only the proposed method gives perfect performances on the three cubes synthetic data and on the Pavia U data set.}
\end{table*}

\subsection{Estimation of Number of Clusters}
\label{subsec:EstimatingK}
    
\begin{figure}[!htb]
    \centering
    \begin{subfigure}{0.11\textwidth}
        \includegraphics[width=\textwidth]{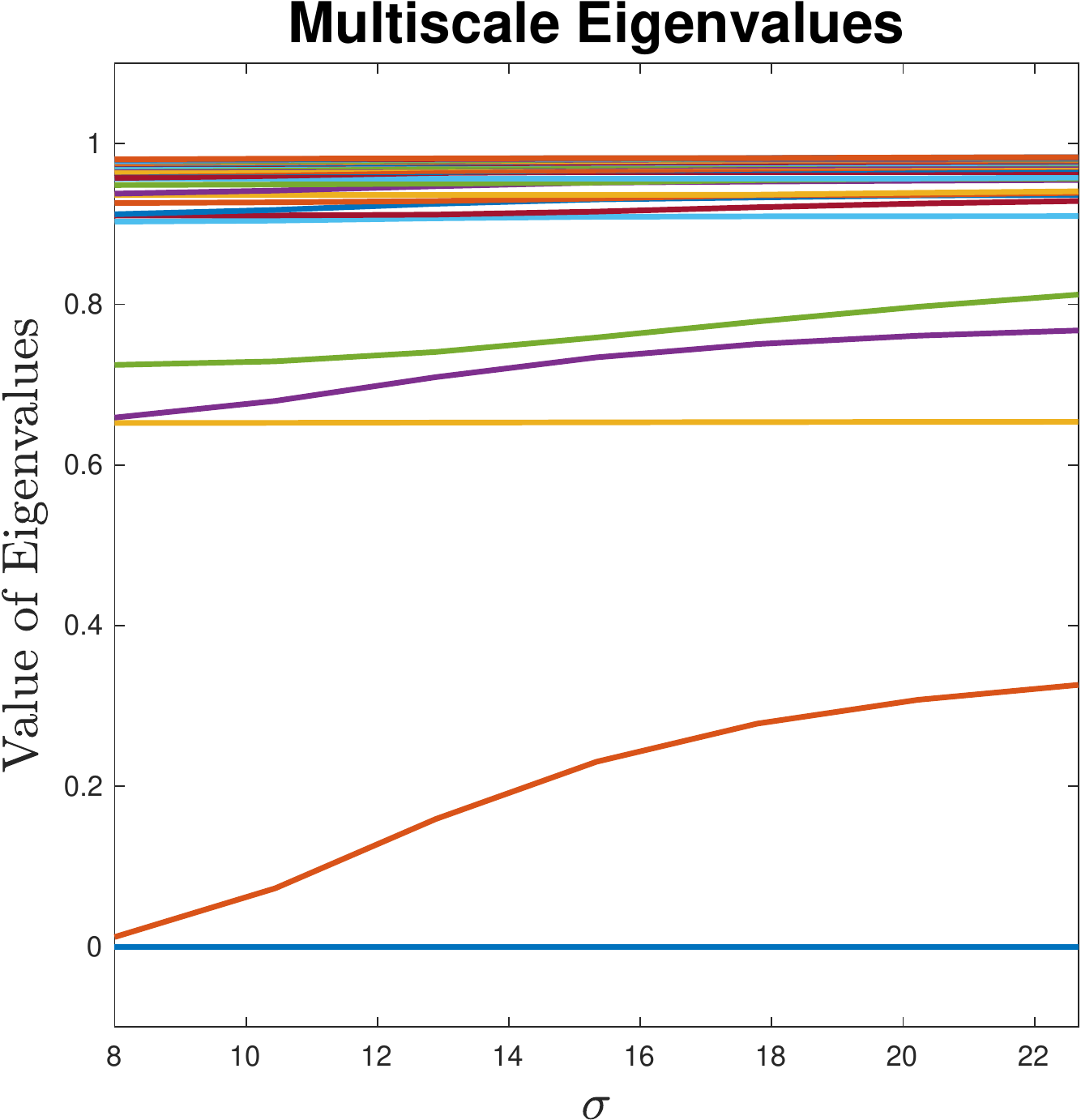}
        \subcaption{FS}
    \end{subfigure}
    \begin{subfigure}{0.11\textwidth}
        \includegraphics[width=\textwidth]{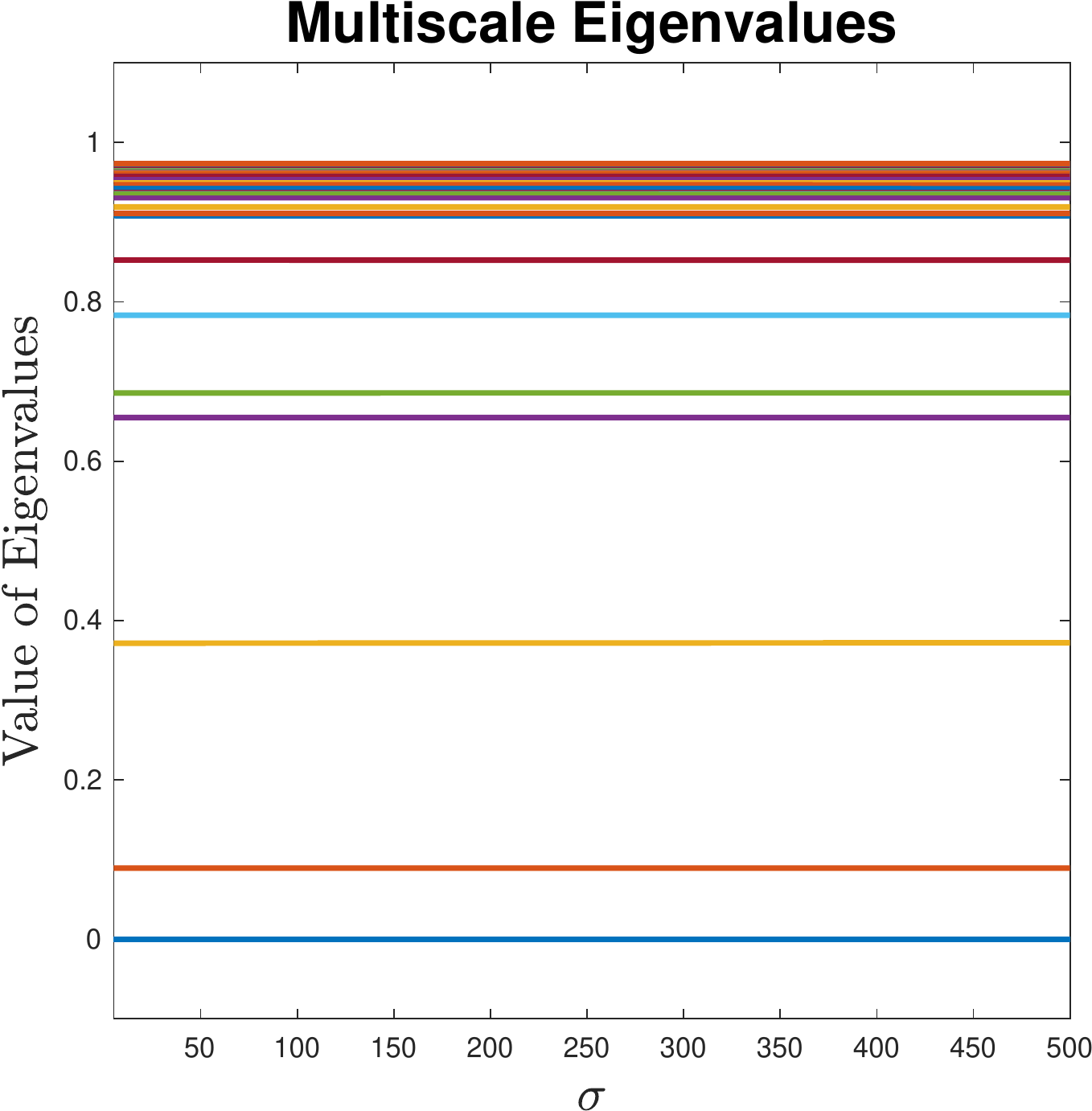}
        \subcaption{TC}
    \end{subfigure}
    \begin{subfigure}{0.11\textwidth}
        \includegraphics[width=\textwidth]{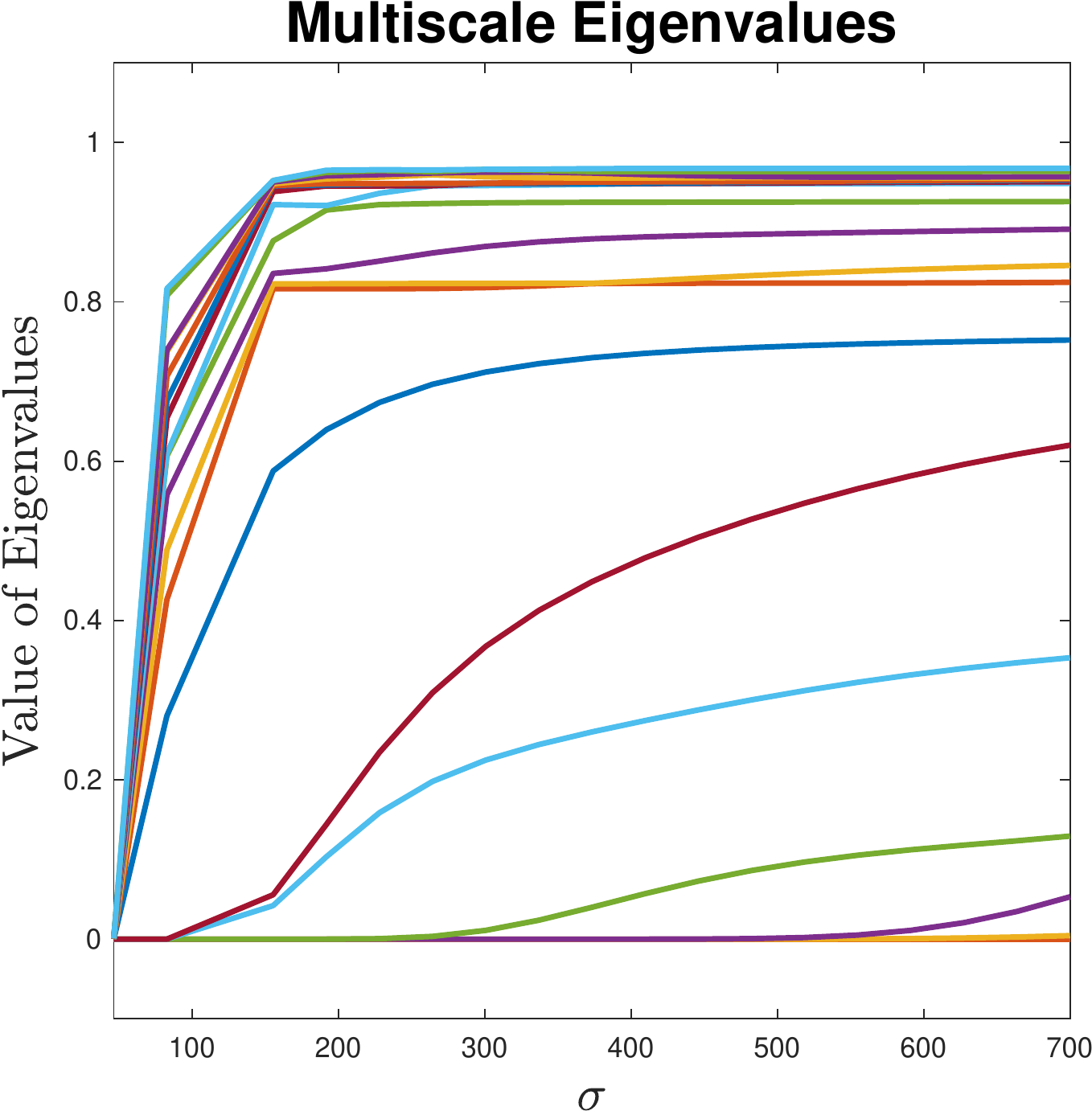}
        \subcaption{Salinas A}
    \end{subfigure}
    \begin{subfigure}{0.11\textwidth}
        \includegraphics[width=\textwidth]{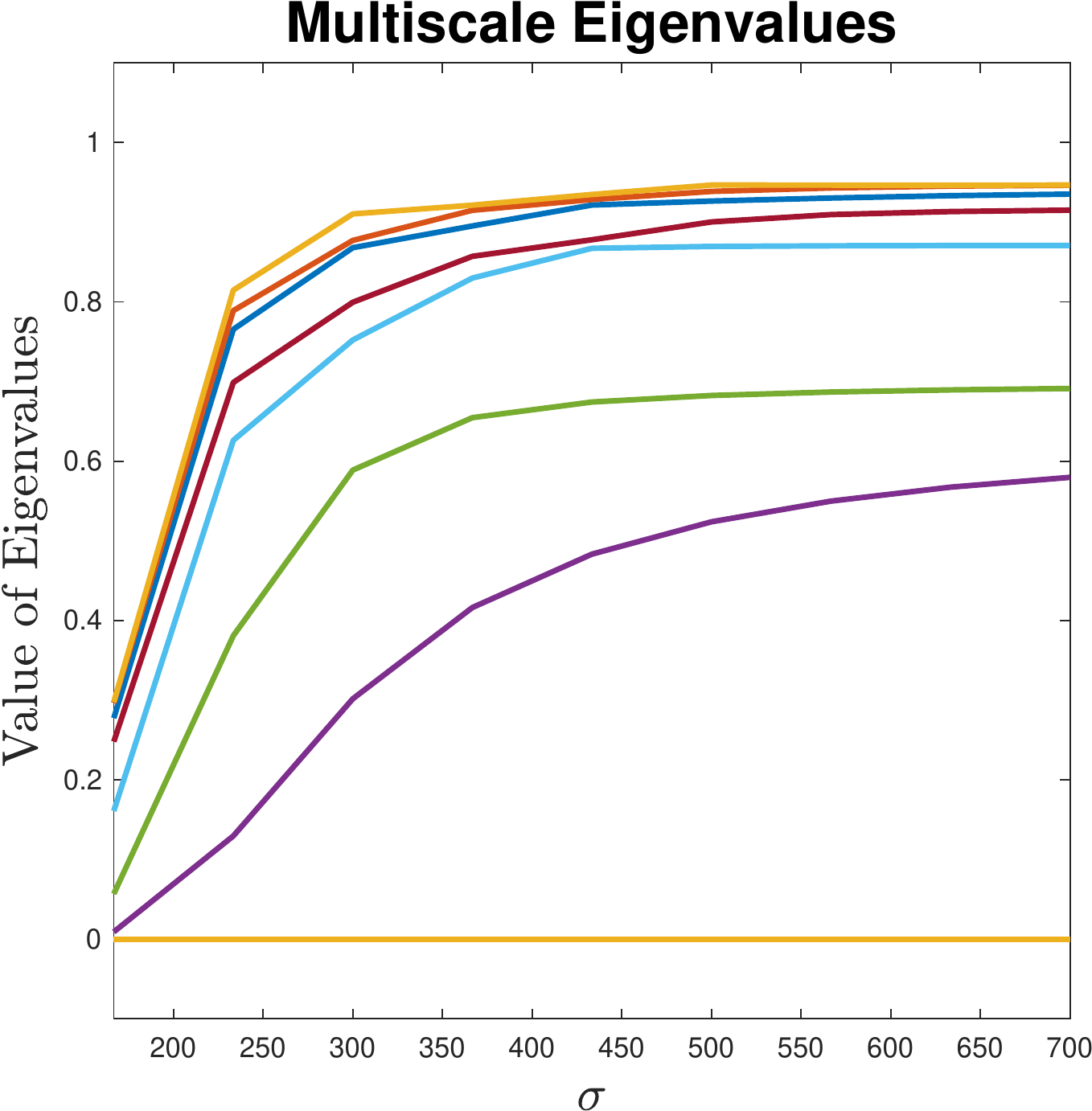} 
        \subcaption{Pavia U}
    \end{subfigure}
    \caption{\label{fig:MultiscaleEigenvalues}For each of the data sets, eigenvalues are shown as a function of $\sigma$.  The first eigenvalue (blue) is always 0; the second eigenvalue is red, the third yellow, and so on.  We see that for a range of $\sigma$ values, the largest gap is between the $K^{th}$ and $(K+1)^{st}$ eigenvalues for both synthetic data sets, as well as Pavia U, indicating that SRUSC is correctly estimates $K$ for these data.  The proposed method estimates $K=7$ for Salinas A, rather than 6.}
    \end{figure}

In Fig. \ref{fig:MultiscaleEigenvalues}, the eigenvalues of the SRUSC Laplacian are shown for a range of scales $\sigma$.  We see the eigengap correctly estimates the number of clusters for most of the data sets considered, for a range of $\sigma$ values.  This suggests that SRUSC is able to estimate the number of clusters even on challenging, high-dimensional HSI.  The multiscale eigenvalues were also computed with the Euclidean Laplacian, with very poor results.
 
\subsection{Runtime}
    
The runtimes for all algorithms are in Table \ref{tab:runtime}.  All experiments were performed on a Macbook Pro with a 3.5GHz Intel Core i7 processor and 16GB RAM.  The runtimes for SC and SRUSC are much higher than for the other comparison methods because these methods perform model selection by using the eigengap statistic to estimate the number of clusters.  This requires running the algorithm over many instances of scaling parameter $\sigma$.  If the number of clusters is know a priori, then the run time for SC and SRUSC are comparable to DL.  

\begin{table}[htb!]
\centering
\begin{adjustbox}{max width=.5\textwidth}
\begin{tabular}{ | c | c | c | c | c |  }
 \hline
 Data set & FS  & TC & SalinasA & PaviaU \\
 \hline
KM  & 0.1072    & 0.0654       & 0.3206      & 0.1590   \\
PCA     & 0.0558    & 0.0514       & 0.0788      & 0.0479    \\
GMM     & 0.2590    & 0.2693       &  1.3530     & 0.1662    \\
SC      & 22.8741   & 39.0727     & 310.1795    & 6.8315   \\
DL      & 1.4789    & 2.4084       & 13.3009.    & 1.2118    \\
FSFDPC  & 1.0794    & 1.5634       & 9.6928      & 0.4860    \\
NMF     & 0.1130    & 0.2789       & 0.6593      & 0.3272    \\ 
LCMR    & 12.3922   & 13.9870     & 35.2812     & 3.5041    \\
SRUSC     & 37.7156   & 39.2982     & 975.8000    & 22.9041   \\
 \hline
\end{tabular}
\end{adjustbox}
\caption{\label{tab:runtime}Running time for different methods.}
\end{table}
\section{Conclusions and Future Research}
\label{sec:Conclusions}

In this article, we showed that UPD are a powerful and efficient metric for the unsupervised analysis of HSI.  When embedded in the spectral clustering framework and combined with suitable spatial regularization, state-of-the-art clustering performance is realized on a range of synthetic and real data sets.  Moreover, ultrametric spectral clustering is mathematically rigorous and enjoys theoretical understanding that many unsupervised learning algorithms lack.

Based on the success of unsupervised learning with ultrametric path distances, it is of interest to develop semisupervised path distance approaches for HSI.  Path distances for semisupervised learning are effective in several contexts \cite{Bijral2011_Semi}, and it is of interest to understand how the UPD perform specifically in the context of high-dimensional HSI,  specifically for \emph{active learning} of HSI \cite{Murphy2019_Learning, Murphy2019_Spatially_1}.  

\section*{Acknowledgements}  This research is partially supported by the US National Science Foundation grants DMS-1912737, DMS-1924513, and CCF-1934553.

   \bibliographystyle{unsrt}
\bibliography{GRSL.bib}
\end{document}